%% file: paper.tex
\documentclass{article}

\usepackage[preprint]{neurips_2025}

\usepackage[utf8]{inputenc} % allow utf-8 input
\usepackage[T1]{fontenc}    % use 8-bit T1 fonts
\usepackage{hyperref}       % hyperlinks
\usepackage{url}            % simple URL typesetting
\usepackage{booktabs}       % professional-quality tables
\usepackage{amsfonts}       % blackboard math symbols
\usepackage{nicefrac}       % compact symbols for 1/2, etc.
\usepackage{microtype}      % microtypography
\usepackage[pdftex]{graphicx}
\usepackage[export]{adjustbox} % for border-radius and frame options in \includegraphics

\usepackage{arydshln}

\usepackage[table]{xcolor}
\definecolor{best}{RGB}{210, 230, 250} % Very light blue (darker than blue!10)
\definecolor{second}{RGB}{235, 245, 255} % Extremely light blue (lighter than best)
\newcommand{\best}[1]{\cellcolor{best}\textbf{#1}}
\newcommand{\second}[1]{\cellcolor{second}\textbf{#1}}

\newcommand{\bestnt}[1]{\colorbox{best}{\textbf{#1}}}
\newcommand{\secondnt}[1]{\colorbox{second}{\textbf{#1}}}

\usepackage{multirow}

\usepackage{subcaption}
\usepackage{adjustbox}
\usepackage{tcolorbox}

\usepackage{amsmath}

\title{From Generation to Generalization: Emergent Few-Shot Learning in Video Diffusion Models}

\author{%
  Pablo Acuaviva\thanks{Computer Vision Group, University of Bern — \url{https://www.cvg.unibe.ch/people/}} \\
  \texttt{pablo.acuavivahuertos@unibe.ch} \\
  \And
  Aram Davtyan\footnotemark[1] \\
  \texttt{aram.davtyan@unibe.ch} \\
  \And
  Mariam Hassan\thanks{VITA Lab, EPFL — \url{https://www.epfl.ch/labs/vita/people/}} \\
  \texttt{mariam.hassan@epfl.ch} \\
  \And
  Sebastian Stapf\footnotemark[1] \\
  \texttt{sebastian.stapf@unibe.ch} \\
  \And
  Ahmad Rahimi\footnotemark[2] \\
  \texttt{ahmad.rahimi@epfl.ch} \\
  \And
  Alexandre Alahi\footnotemark[2] \\
  \texttt{alexandre.alahi@epfl.ch} \\
  \And
  Paolo Favaro\footnotemark[1] \\
  \texttt{paolo.favaro@unibe.ch} \\
}

\begin{document}

\maketitle

\begin{abstract}
Video Diffusion Models (VDMs) have emerged as powerful generative tools, capable of synthesizing high-quality spatiotemporal content. Yet, their potential goes far beyond mere video generation. We argue that the training dynamics of VDMs, driven by the need to model coherent sequences, naturally pushes them to internalize structured representations and an implicit understanding of the visual world. To probe the extent of this internal knowledge, we introduce a few-shot fine-tuning framework that repurposes VDMs for new tasks using only a handful of examples. Our method transforms each task into a visual transition, enabling the training of LoRA weights on short input–output sequences without altering the generative interface of a frozen VDM. Despite minimal supervision, the model exhibits strong generalization across diverse tasks, from low-level vision (for example, segmentation and pose estimation) to high-level reasoning (for example, on ARC-AGI). These results reframe VDMs as more than generative engines. They are adaptable visual learners with the potential to serve as the backbone for future foundation models in vision. See our project page: \url{https://pabloacuaviva.github.io/Gen2Gen/}.
\end{abstract}

\section{Introduction}
\begin{quote}
\emph{“What I cannot create, I do not understand.”} \\
\hfill --- Richard P. Feynman (see \cite{feynman1989})
\end{quote}

Generative AI has taken the world by storm. Over the past few years, we have witnessed astonishing advances in the ability of machines to synthesize coherent text \cite{achiam2023gpt}, images \cite{flux2024}, audio \cite{sesame_uncanny_valley_voice, nari2025dia}, and even video \cite{polyak2025moviegencastmedia}. These developments are largely driven by two dominant paradigms in generative modeling: Large Language Models (LLMs) \cite{achiam2023gpt, grattafiori2024llama, geminiteam2024gemini15, deepseekr1_2025}, trained via next-token prediction on vast corpora of text, and diffusion models \cite{ho2020denoising, rombach2022highresolutionimagesynthesislatent}, which approach generation by progressively denoising data from random noise.
% and which are built on transformer architectures \cite{Vaswani2017AttentionIA}

A defining characteristic of LLMs is their potential at few-shot learning, enabling them to adapt remarkably well to new tasks with in-context examples \cite{brown2020language} or through parameter-efficient fine-tuning (PEFT) \cite{liu2022fewshotparameterefficientfinetuningbetter} on a few samples. The remarkable ability of LLMs for rapid adaptation via prompting and efficient specialization through PEFT methods has motivated exploration into analogous generalist models for images \cite{wang2023images, xiao2024omnigen, lin2025realgeneral}. A central challenge in the image domain is realizing powerful vision models that exhibit similar versatility across a spectrum of visual tasks, guided by contextual cues and limited examples \cite{bai2024sequential, yang2024video}.

An especially promising frontier is Video Diffusion Models (VDMs) \cite{blattmann2023stablevideodiffusionscaling, zheng2024opensorademocratizingefficientvideo, polyak2025moviegencastmedia, brooks2024video, deepmind2024veo2}, which extend generative modeling into the spatiotemporal domain. Unlike static image generation, video synthesis requires models to capture both high-fidelity spatial details within individual frames and consistent temporal dynamics across sequences.
To generate coherent video, VDMs must construct an implicit generative model of visual data, continuously minimizing the discrepancy between their predictions and the actual data, a process analogous to the theorized minimization of free energy in biological systems \cite{parr2022active, friston2010free}. We argue that \textbf{this process incentivizes VDMs to internalize rich, structured representations and acquire an implicit understanding of visual content}. 

To validate this hypothesis, we propose a \textbf{few-shot fine-tuning} approach, in which a VDM is adapted to new tasks using only a small number of training examples. The key intuition is that successful adaptation under such constraints implies the model already encodes the relevant knowledge -- thus revealing the scope of its internal representations and prior capabilities.
% \footnote{In our experiments, the pre-trained VDMs are also text-conditioned; however, we fix the text input to a neutral prompt to ensure it does not influence the evaluation.}

Our fine-tuning framework reformats training examples to be compatible with the VDM. For each input–output image pair specifying a task, we construct a short video depicting the transition from input to output, which serves as the target sequence for the model to generate when conditioned on the input image.

Using this strategy, we demonstrate that VDMs can be repurposed to solve a broad spectrum of vision tasks, from classical computer vision problems to abstract reasoning, well beyond their original training objectives. Remarkably, even when fine-tuned on only a few such examples, VDMs exhibit strong generalization capabilities. These findings suggest that VDMs are not merely generative tools but highly adaptable visual learners. Their ability to generalize from minimal supervision points toward a future of visual foundation models that are grounded in generation and capable of rapid adaptation. This represents an essential step toward building intelligent and flexible vision systems.

\begin{tcolorbox}[colback=second, colframe=second, width=\textwidth, boxrule=0pt, arc=0pt]
\textbf{Summary of contributions.}
\begin{enumerate}
    \item We propose a unified fine-tuning framework for adapting pre-trained VDMs to diverse visual tasks in a few-shot setting;
    
    \item Through extensive experiments on a wide range of visual tasks, we demonstrate the emergent latent capabilities of pre-trained VDMs;
    
    \item To the best of our knowledge, we are the first to successfully apply VDMs to visual reasoning tasks on the Abstract Reasoning Corpus (ARC-AGI) benchmark \cite{chollet2019measure}, significantly expanding their demonstrated reasoning abilities.
\end{enumerate}
\end{tcolorbox}

\section{Related Work}

\noindent\textbf{Few-shot learning} LLMs have demonstrated strong few-shot capabilities \cite{brown2020language}, both in-context \cite{dong2024surveyincontextlearning} and through PEFT approaches \cite{liu2022fewshotparameterefficientfinetuningbetter, najafi2024rifflearningrephraseinputs}. While few-shot fine-tuning has also been applied to text-to-image (T2I) diffusion models \cite{giannone2022few}, most studies focus on personalization tasks \cite{ruiz2023dreambooth, liu2024ada}. Current approaches instead tend to make use of diffusion model features \cite{tang2023emergent, helbling2025conceptattention, delatolas2025studying, wang2024zero} for tackling few-shot or zero-shot tasks.\\
\noindent\textbf{Video Diffusion Models} The field of video generation has seen significant advancements through the application of diffusion models, particularly in text-to-video (T2V) and image-to-video (I2V) generation. Early approaches, such as CogVideo \cite{hong2022cogvideolargescalepretrainingtexttovideo} and Phenaki \cite{villegas2022phenakivariablelengthvideo}, focused on scaling Transformers \cite{Vaswani2017AttentionIA} for video generation. More recently, diffusion-based methods have shown substantial progress, with models like Sora \cite{brooks2024video}, MovieGen \cite{polyak2025moviegencastmedia}, Veo 2 \cite{deepmind2024veo2} and CogVideoX \cite{yang2024cogvideox} setting new benchmarks in quality and realism. Alongside these, LTX-Video \cite{hacohen2024ltx} distinguishes itself with exceptionally fast generation speeds, offering real-time generation capabilities.
Recent research has also explored controllable video generation \cite{agarwal2025cosmos, hassan2024gemgeneralizableegovisionmultimodal, kanervisto2025world}, aiming to achieve high-quality, realistic video rendering with fine-tuned control over motion and dynamics. These approaches focus on inferring dynamic environments and predicting plausible future states from past observations and control signals.\\
\noindent\textbf{Generalist Vision Models}
Recent work has shown that generative models can be trained to act as generalist vision models. Image inpainting as visual prompting \cite{bar2022visual} and image-based in-context learning \cite{wang2023images} highlight the capacity of generative models to solve diverse tasks via structured inputs. Diffusion models have been further adapted for in-context learning \cite{wang2023context} and instruction-following across tasks \cite{geng2024instructdiffusion}. Sequential modeling has also been proposed as a unified training interface for scaling vision models \cite{bai2024sequential}. Closest to our work is RealGeneral \cite{lin2025realgeneral}, which fine-tunes CogVideoX1.5 for multi-task learning via temporal in-context prompts. Unlike their full fine-tuning approach with abundant data, we apply LoRA-based few-shot tuning for each task by creating videos that help align tasks with the model inductive biases, enabling more sample-efficient generalization.

\begin{figure}[t]
  \centering
  \includegraphics[width=1.0\linewidth]{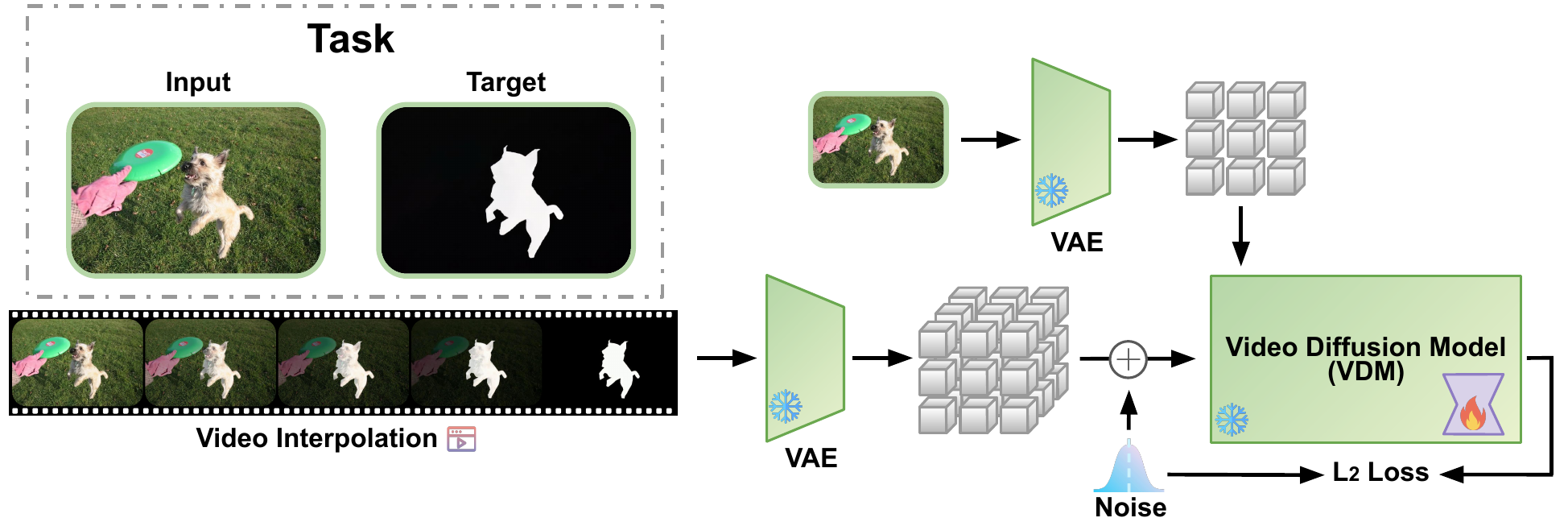}
  \caption{
    Proposed framework. Given a task encoded as input-target image pairs (dashed gray box), a transition video is constructed to transform the input into the target. We fine-tune LoRA adapters with the core model frozen. At inference, the model outputs a transition video from a new input, with the final frame used as the prediction.
    }
  \label{fig:method}
\end{figure}

\section{Probing Latent Knowledge in VDMs with Few-Shot Learning}

\subsection{Mapping Image-Based Tasks to Videos}
% Do we want to include that much detail in the problem formulation? Or just provide a quick overview? 
% Then in 3.2 Proposed approach we can give more details about specifics on each of the sections? 
We consider a task \(\mathcal{T}\) represented by a dataset \(\mathcal{D}_{\mathcal{T}} = \{(x_i, y_i)\}_{i=1}^{n}\), where \(x_i, y_i \in \mathbb{R}^{3 \times H \times W}\) are pairs of input-output RGB images with height \(H\) and width \(W\).
The task \(\mathcal{T}\) consists of learning a mapping that transforms any input image \(x\) to its corresponding output \(y\). For instance, \(x\) could be a natural image and \(y\) the segmentation mask \(x\) (see Figure~\ref{fig:method}). In our few-shot setting, the number \(n\) of such pairs is quite small, typically in the range of \(n=3\) to \(n=30\). We propose to learn this mapping via a VDM by creating a dataset \(\mathcal{V}_{\mathcal{T}}=\{[x_i, v_{i,2}, \dots, v_{i,F-1}, y_i]\}_{i=1}^n\) of artificial sequences of \(F\) frames that are transitions between \(x_i\) and \(y_i\), where \(x_i\) and \(y_i\) are pairs from the dataset \(\mathcal{D}_{\mathcal{T}}\).

This dataset is then directly compatible with pre-trained VDMs that generate video sequences conditioned on a single input frame. For each sequence \(v = [v_1, v_2, \dots, v_{F-1}, v_F] \in \mathcal{V}_\mathcal{T}\), we fine-tune a VDM to denoise all the frames \([v_1,\dots, v_F]\) by using \(v_1\) as a conditioning input. At inference, the fine-tuned VDM can be used to generate a video sequence \(\bar v = [x, \bar v_2,\dots, \bar{v}_F]\) by using a new image \(x\) as conditioning. We then obtain the predicted output \(\bar y = \bar v_F\) corresponding to \(x\) by extracting the last frame in the generated video.

In the following section, we briefly introduce the key background underlying our approach before describing the method in detail.

\subsection{Background}

\paragraph{Video Diffusion Models}  
VDMs extend denoising diffusion probabilistic models \cite{ho2020denoising} to the temporal domain \cite{blattmann2023stablevideodiffusionscaling}. The forward process adds Gaussian noise to video frames \( v^0 \sim q(v) \) over time and is described by the forward process distributions
\begin{equation}
q(v^t \mid v^{t-1}) = \mathcal{N}(v^t; \sqrt{1 - \beta_t} v^{t-1}, \beta_t \mathbf{I}),
\end{equation}
where $\beta_t$ is the variance schedule with diffusion step  $t\in[1,T]$, and $T$ is the total number of diffusion steps \cite{ho2020denoising}. In our notation, the distribution $q$ is represented by our video dataset $\mathcal{V}_\mathcal{T}$.
A VDM is then a neural network \( \epsilon_\theta \) with parameters (weights) $\theta$, which predicts noise to obtain the reverse process
\begin{equation}
p_\theta(v^{t-1} \mid v^t) = \mathcal{N}(v^{t-1}; \mu_\theta(v^t, t, c), \sigma_t^2 \mathbf{I}),
\end{equation}
where the variance $\sigma_t^2$ is a function of $\beta_t$ and $\mu_\theta(v^t, t, c)$ is a function of $v^t$ and $\epsilon_\theta$, and where \( c \) denotes conditioning information. In our case, $c=\{v^0_1,e_{\text{text}}\}$, where $e_{\text{text}}$ is a text prompt embedding. In our analysis, the text prompt $e_{\text{text}}$ is fixed to a neutral (task-agnostic) prompt across all fine-tunings and evaluations.
$\epsilon_\theta$ is trained by minimizing the objective

\begin{equation}
\mathcal{L}(\theta) = \mathbb{E}_{v^0\sim \mathcal{V}_\mathcal{T}, \epsilon\sim \mathcal{N}(0, \mathbf{I}), t\sim [1,\dots, T]} \left[ \| \epsilon - \epsilon_\theta(v^t, t, c) \|_2^2 \right].
\label{eq:denoising}
\end{equation}

\paragraph{Low-Rank Adaptation}  
To probe the existing knowledge in a pre-trained VDM, we freeze all model weights \( \theta = \{ \theta_{\ell} \}_{\ell=1}^L \), where \(L\) is the number of model layers, and fine-tune low-rank adaptation (LoRA) parameters \cite{hu2021loralowrankadaptationlarge} on top of them. This procedure consists of augmenting selected weight matrices \( \theta_{\ell} \in \mathbb{R}^{d \times k} \) for $\ell\in \Omega\subset [1,\dots,L]$ with trainable low-rank updates
\begin{equation}
\hat{\theta}_{\ell} =  \theta_{\ell} + \Delta  \theta_{\ell} = \theta_{\ell} + B_\ell A_\ell,
\end{equation}
where \( B_\ell \in \mathbb{R}^{d \times r} \), \( A_\ell \in \mathbb{R}^{r \times k} \), and \( r \ll \min(d, k) \). Fine-tuning minimizes \( \mathcal{L}(\{\hat{\theta}_\ell\}_{\ell\in\Omega}) \) in Eq.~\eqref{eq:denoising} with respect to the LoRA parameters \( \{A_\ell, B_\ell\} \), while keeping all \( \theta_{\ell} \) fixed.

\begin{figure}[t]
  \centering
  \begin{subfigure}[b]{0.13\textwidth}
    \includegraphics[width=\textwidth]{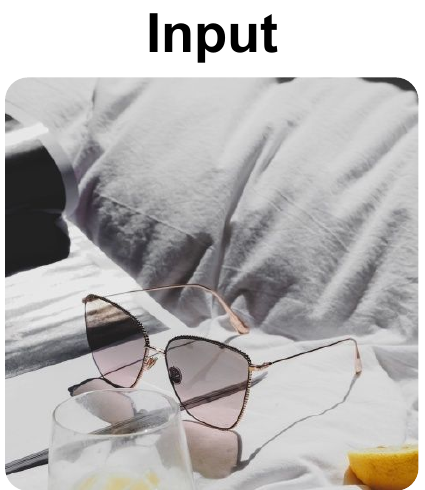}
    \subcaption{}
  \end{subfigure}
  \hfill
  \begin{subfigure}[b]{0.13\textwidth}
    \includegraphics[width=\textwidth]{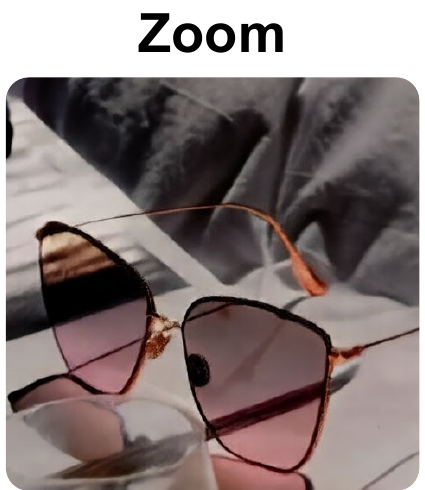}
    \subcaption{}
  \end{subfigure}
  \hfill
  \begin{subfigure}[b]{0.13\textwidth}
    \includegraphics[width=\textwidth]{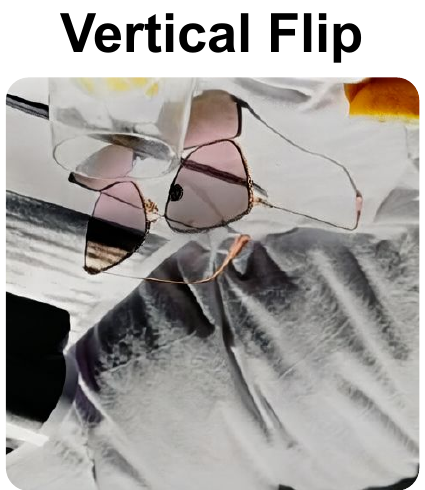}
    \subcaption{}
  \end{subfigure}
  \hfill
  \begin{subfigure}[b]{0.13\textwidth}
    \includegraphics[width=\textwidth]{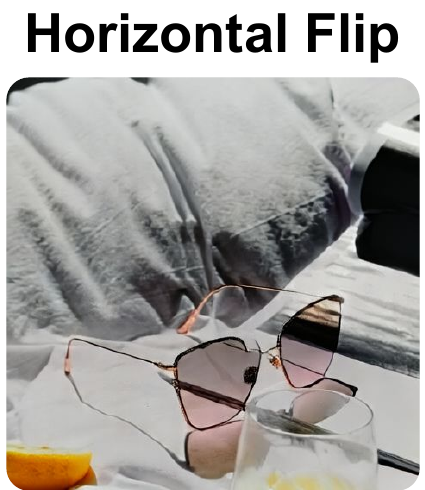}
    \subcaption{}
  \end{subfigure}
  \hfill
  \begin{subfigure}[b]{0.13\textwidth}
    \includegraphics[width=\textwidth]{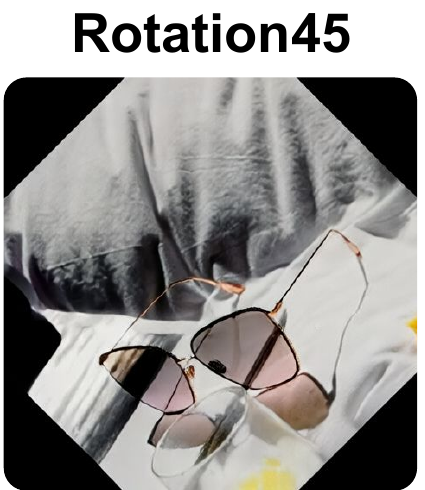}
    \subcaption{}
  \end{subfigure}
  \hfill
  \begin{subfigure}[b]{0.13\textwidth}
    \includegraphics[width=\textwidth]{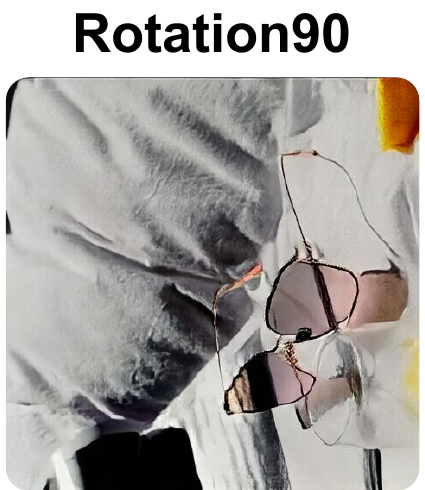}
    \subcaption{}
  \end{subfigure}
  \hfill
  \begin{subfigure}[b]{0.13\textwidth}
    \includegraphics[width=\textwidth]{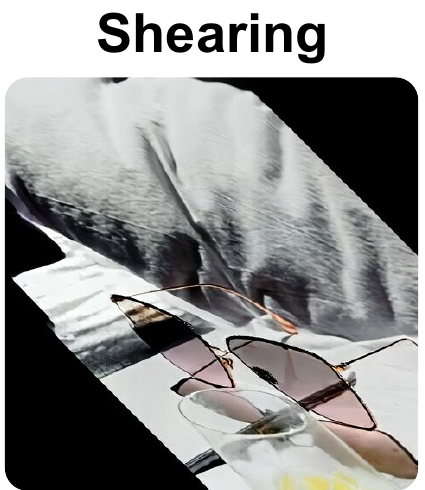}
    \subcaption{}
  \end{subfigure}
  \caption{Examples of learned transformations applied to a new image (a) in 1-shot setting (b), (c), and (d), and 3-shot setting (e), (f), and (g).}
  \label{fig:misc/image-transforms}
\end{figure}

\begin{figure}[t]
  \centering
  \includegraphics[width=1.0\linewidth]{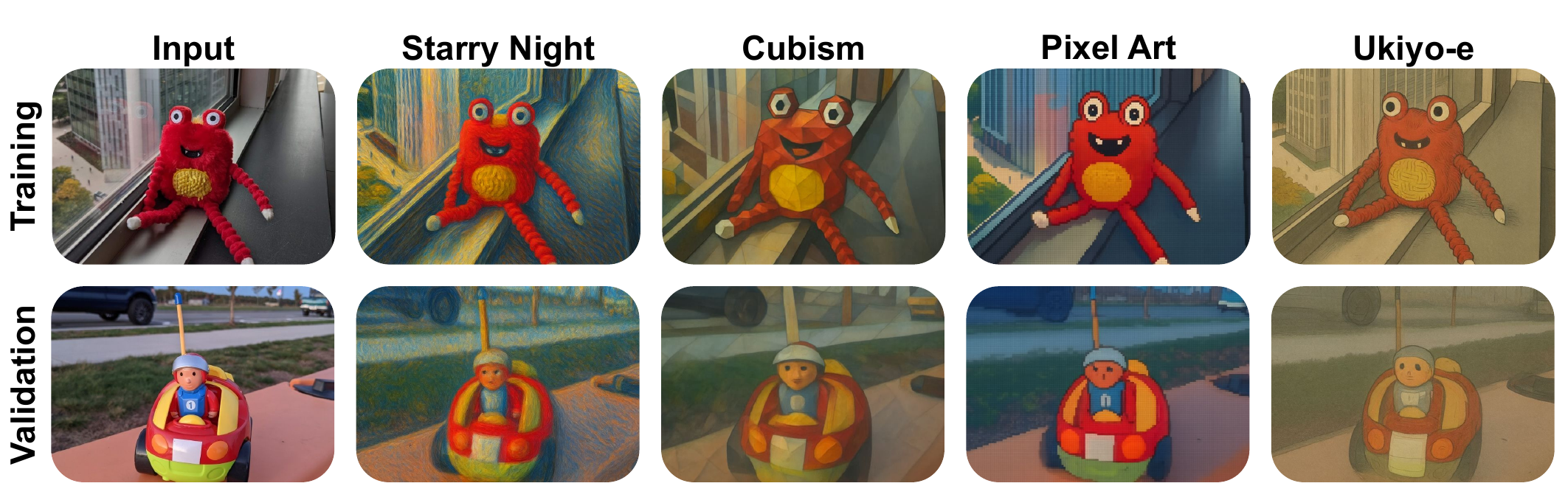}
  \caption{Style transfer in a 1-shot setting, applied to a new image (second row).}
  \label{fig:misc/style-transfer-mini}
\end{figure}

\begin{figure}[t]
  \centering  \includegraphics[width=0.75\linewidth]{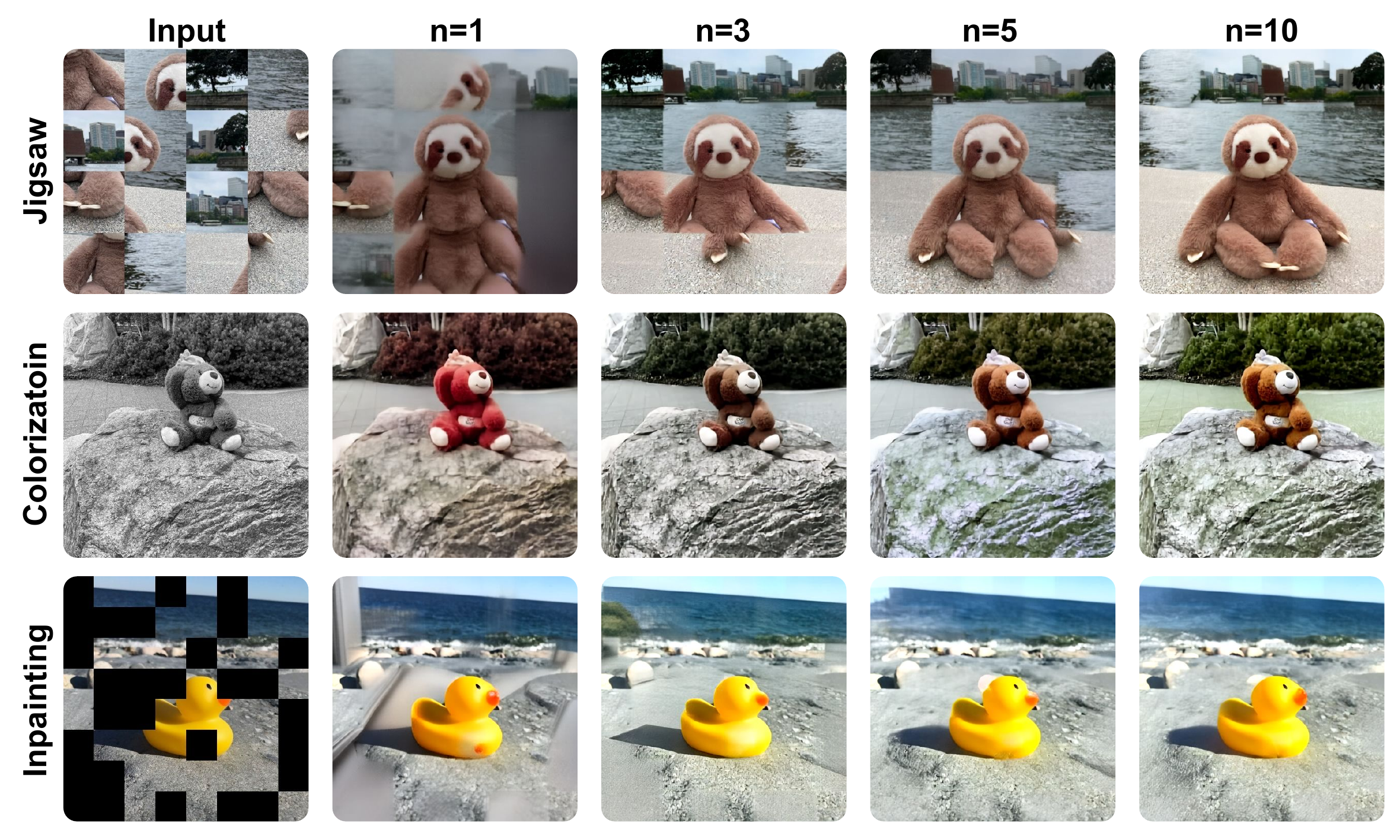}
  \caption{Qualitative performance on different vision tasks with an increasing number of training samples from \(n=1\) to \(n=10\).}
  \label{fig:misc/color-jigsaw-inpaint}
\end{figure}

\subsection{Few-Shot Fine-Tuning and Inference}\label{sec:few-shot-framework}

Our framework to adapt VDMs to novel visual tasks is split into three steps.

\paragraph{1. Video Trajectory Interpolation}
We introduce a parameterized video interpolation function \(\phi:\mathbb{R}^{3 \times H \times W} \times \mathbb{R}^{3 \times H \times W} \times [1,F] \rightarrow \mathbb{R}^{3 \times H \times W}\) to model the transformation between input \(x\) and target image \(y\) as a video sequence \(v = [v_1, \ldots, v_F]\)

such that \(v_f = \phi(x,y,f)\), for \(f=[1,\dots, F]\). For example, a simple choice is the linear combination 
\begin{align}
    \phi(x,y,f) = \left(1-\frac{f-1}{F-1}\right)x + \frac{f-1}{F-1}y
\end{align}
such that \(v_1 = \phi(x,y,1) = x\) and \(v_F = \phi(x,y,F) = y\).
The video interpolation function is then used to create the video dataset $\mathcal{V}_\mathcal{T}=\{[\phi(x_i,y_i,1),\dots,\phi(x_i,y_i,F)]\}_{i=1}^n$. 

\paragraph{2. Task-Specific Adaptation}
We use a pretrained image-to-video (I2V) diffusion model as our VDM. To adapt the model to a specific task \(\mathcal{T}\), we fine-tune LoRA weights in different layers inside the model. For training we use the dataset $\mathcal{V}_\mathcal{T}$. The training procedure is illustrated in Figure~\ref{fig:method}.

\paragraph{3. Inference Procedure} Given a new test image $x_{\text{test}}$, we generate the target image as follows:
\begin{enumerate}
\item Construct conditioning vector $c_{\text{test}} = \{x_{\text{test}}, e_{\text{text}}\}$.
\item Sample initial noise $v^{T} \sim \mathcal{N}(\mathbf{0}, \mathbf{I})$.
\item Apply denoising for $t = T, \ldots, 1$ to $v^t$.
\item Extract the final frame $v^0_F$ from the denoised sequence as our prediction $\hat{y}$.
\end{enumerate}

By reframing the task as a video generation problem, our approach naturally aligns with the inductive biases of video diffusion models. By leveraging lightweight LoRA modules, we steer the generative process toward task-specific behavior with minimal supervision. This combination enables efficient few-shot adaptation while preserving most of the original knowledge in the VDM.

\section{Experiments}

We conduct a variety of experiments to evaluate our proposed framework. Unless otherwise specified, all experiments use the I2V version of CogVideoX1.5~\cite{yang2024cogvideox}. 

\subsection{Initial Exploration}\label{sec:initial_exploration}

We begin by demonstrating that our framework can steer the model to learn geometric transformations in a few-shot setting (1 or 3-shot), as shown in Figure~\ref{fig:misc/image-transforms}. For these tasks, we use the dataset provided in \cite{ruiz2023dreambooth} and apply the corresponding transformations. In these few-shot scenarios, the model’s ability to effectively generalize suggests that it must rely on strong, well-aligned priors. These priors enable the model to produce the desired transformations from minimal data, with the LoRAs helping to steer the behavior efficiently. In a similar spirit, we show that we can also steer the model to handle 1-shot style transfer tasks, see Figure~\ref{fig:misc/style-transfer-mini}.

For more challenging tasks, such as Jigsaw puzzles, colorization and inpainting, in Figure~\ref{fig:misc/color-jigsaw-inpaint} we qualitatively show the impact of an increasing number of training samples on the  performance. 

\begin{figure}[t]
  \centering
  \begin{subfigure}[b]{0.35\linewidth}
    \centering
    \includegraphics[width=\linewidth]{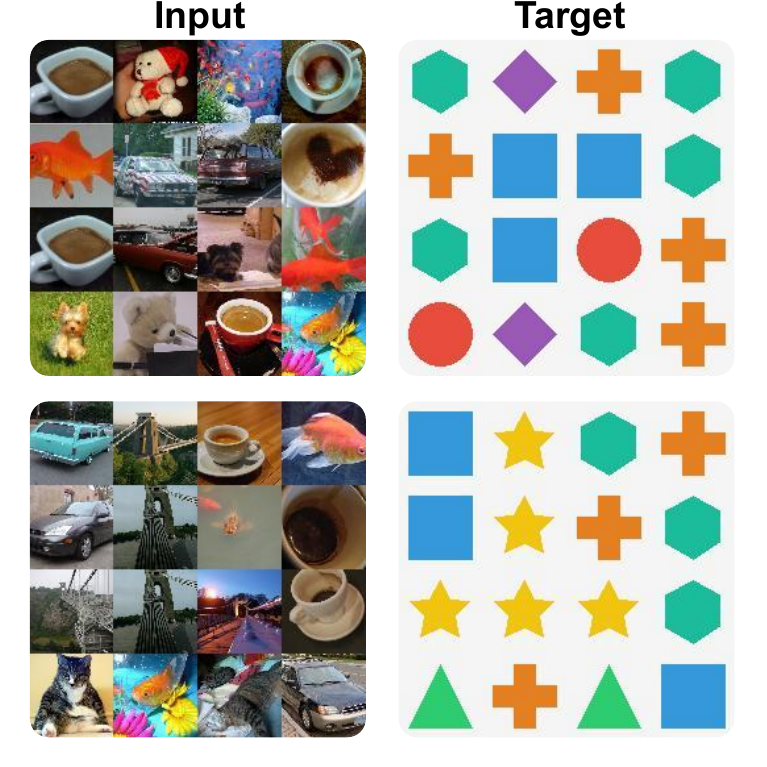}
    \caption{}
    \label{fig:misc/tinyimagenet-classification-a}
  \end{subfigure}
  \hfill
  \begin{subfigure}[b]{0.55\linewidth}
    \centering
    \includegraphics[width=\linewidth]{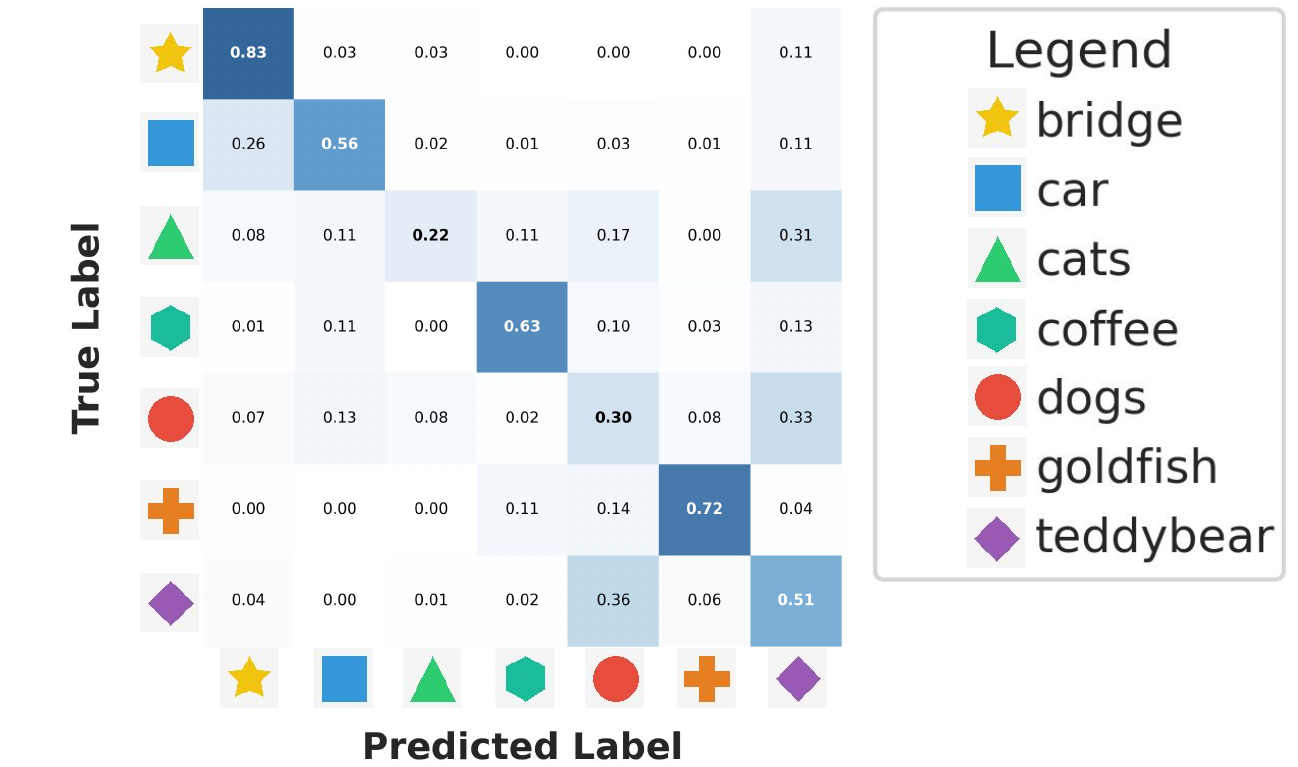}
    \caption{}
    \label{fig:misc/tinyimagenet-classification-b}
  \end{subfigure}
  \caption{Grid classification on TinyImageNet: (a) Training examples, (b) Confusion matrix with \(n=30\) training samples. }
  \label{fig:misc/tinyimagenet-classification}
\end{figure}

Finally, we evaluate our model on an entity classification task using TinyImageNet \cite{tiny-imagenet}. We select seven object classes and construct \(4 \times 4\) grids of images as inputs, with each class assigned a unique symbol and color. These symbols are used to generate the corresponding \(4 \times 4\) target grids, see Figure~\ref{fig:misc/tinyimagenet-classification-a}. After training with \(n = 30\) examples, the model learns to distinguish classes that are visually or semantically distinct. We observe that classification remains more challenging for closely related classes (see the confusion matrix in  Figure~\ref{fig:misc/tinyimagenet-classification-b}).

\subsection{Design Choices} \label{sec:design_choices}

\begin{figure}[t]
  \centering
  \begin{subfigure}[b]{0.65\linewidth}
    \centering
    \includegraphics[width=\linewidth]{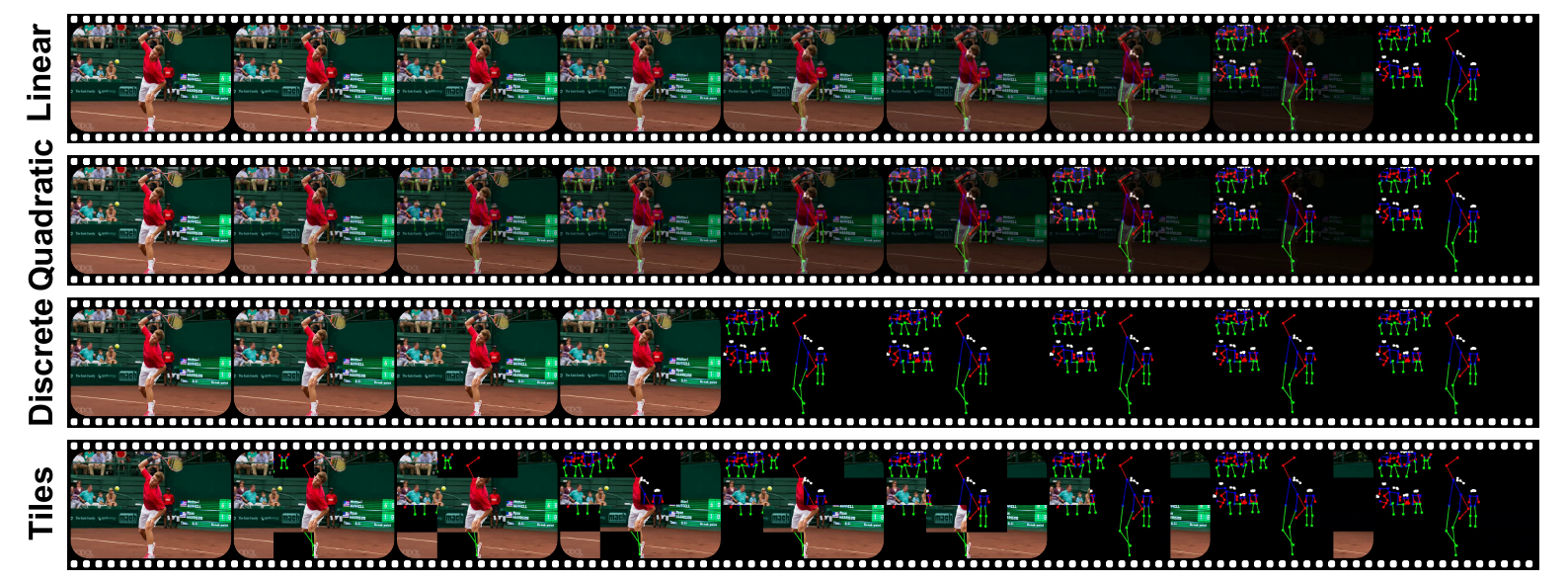}
    \caption{}
    \label{fig:design-choices/interpolation-method-a}
  \end{subfigure}
  \hfill
  \begin{subfigure}[b]{0.33\linewidth}
    \centering
    \includegraphics[width=\linewidth]{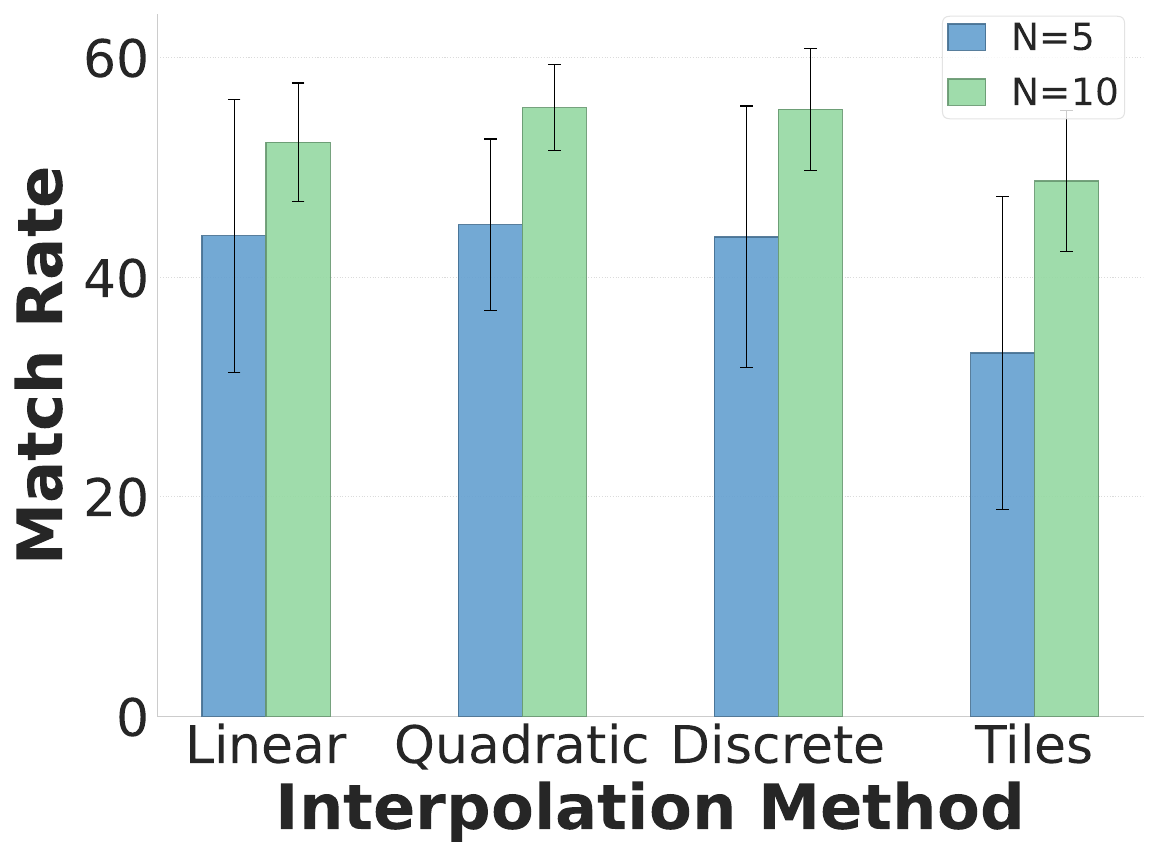}
    \caption{}
    \label{fig:design-choices/interpolation-method-b}
  \end{subfigure}
  \caption{Comparison of different interpolation methods for pose estimation: (a) Interpolation methods examples, (b) Results for pose estimation with \(n=5\) and \(n=10\) training samples.}
  \label{fig:design-choices/interpolation-method}
\end{figure}

We investigate the impact of key design choices in our framework by evaluating it on two representative tasks: binary segmentation and pose estimation. Both evaluations are conducted on selected subsets of the COCO 2017 dataset \cite{lin2014microsoft}.

\begin{itemize}
\item \textbf{Binary Segmentation.} We assess performance on the dog class using the mean Intersection-over-Union (mIoU) metric.
\item \textbf{Pose Estimation.} We propose a custom metric, \textbf{Match Rate}, which quantifies how well the model captures four essential body components: head, torso, arms, and legs. Further details and justification are provided in the Appendix.
\end{itemize}

To better understand the role of individual components and hyperparameters, we perform a systematic analysis of the following factors:

\begin{itemize}
    \item \textbf{Video Interpolation Method.} We explore distinct trajectories in image space and its effect on overall performance. For this, we compare four strategies employed for the video generator $\phi$: \emph{Linear}, \emph{Quadratic}, \emph{Discrete}, and \emph{Tiles}, shown in Figure~\ref{fig:design-choices/interpolation-method-a}. 
    \item \textbf{Target Modules.} We vary the placement of LoRA adapters within the model, targeting query (Q), key (K), value (V), and output (O) projection matrices of the attention layers, or all linear layers. This allows us to evaluate how different insertion points affect adaptation capabilities and generalization performance.
    \item \textbf{LoRA Rank.} We conduct a sweep over different rank values for the LoRA adapters to quantify the trade-off between adapter capacity (number of trainable parameters) and the resulting task performance.
\end{itemize}

\begin{table}[t]
  \caption{Comparison of interpolation methods and LoRA strategies across Binary Segmentation (mIoU ↑) and Pose Estimation (Match Rate ↑) tasks, over increasing number of training samples. \bestnt{Best} and \secondnt{second-best} results are highlighted.}
  \label{tab:ablations/combined-results}
  \centering
  \scriptsize
  \renewcommand{\arraystretch}{1.2}
  \setlength{\tabcolsep}{5pt}
  \begin{tabular}{llcccccccc}
    \toprule
    & & \multicolumn{4}{c}{\textbf{Binary Segmentation (mIoU ↑)}} & \multicolumn{4}{c}{\textbf{Pose Estimation (Match Rate ↑)}} \\
    \cmidrule(lr){3-6} \cmidrule(lr){7-10}
    \textbf{Category} & \textbf{Method} & \(\mathbf{n=3}\) & \(\mathbf{n=5}\) & \(\mathbf{n=10}\) & \(\mathbf{n=30}\) & \(\mathbf{n=3}\) & \(\mathbf{n=5}\) & \(\mathbf{n=10}\) & \(\mathbf{n=30}\) \\
    \midrule
    \multirow{4}{*}{Interpolation}
      & Linear        & 16.56 & 30.36 & \second{46.71} & \second{55.96} & 34.13 & 43.80 & 52.77 & 64.63 \\
      & Quadratic     & \best{21.50} & \second{30.50} & \best{48.97} & \best{58.78} & \best{39.15} & \best{44.80} & \best{55.48} & \second{65.58} \\
      & Discrete      & \second{19.65} & \best{30.72} & 46.56 & 55.36 & \second{36.29} & \second{43.69} & \second{57.23} & 65.48 \\
      & Tiles         & 14.85 & 29.95 & 39.96 & 54.68 & 30.73 & 33.13 & 48.77 & \best{67.12} \\
    \midrule
    \multirow{4}{*}{LoRA Modules}
      & QK            & 18.81 & 25.89 & 41.28 & 51.27 & 27.71 & 29.05 & 39.83 & \second{61.83} \\
      & VO            & \best{23.87} & \second{31.90} & \second{45.46} & 55.17 & \best{39.94} & \second{44.36} & \second{54.36} & 59.10 \\
      & QKVO          & \second{21.50} & 30.50 & \best{48.97} & \best{58.78} & \second{39.15} & \best{44.80} & \best{55.48} & \best{65.58} \\
      & All Linear    & 21.42 & \best{33.32} & 42.61 & \second{56.34} & 35.37 & 42.84 & 52.42 & 58.02 \\
    \midrule
    \multirow{5}{*}{LoRA Rank}
      & Rank 16       & 17.62 & 27.46 & 44.52 & 49.32 & 35.88 & 35.03 & 46.25 & \second{59.46} \\
      & Rank 32       & 18.09 & 30.51 & 44.25 & 48.77 & \second{36.88} & 38.87 & 47.63 & 58.95 \\
      & Rank 64       & 21.50 & \best{30.50} & \best{48.97} & \best{58.78} & \best{39.15} & \best{44.80} & \best{55.48} & \best{65.58} \\
      & Rank 128      & \best{24.48} & 30.11 & 41.98 & \second{55.00} & 36.06 & \second{44.72} & 49.00 & 57.66 \\
      & Rank 256      & \second{21.95} & \second{31.14} & \second{43.67} & 54.23 & 33.38 & 40.25 & \second{49.97} & 59.35 \\
    \bottomrule
  \end{tabular}
\end{table}

To help mitigate the inherent variability with respect to the training set in a few-shot setting, we perform several runs and report the average results when training with a small number of samples. Our results are presented in Table \ref{tab:ablations/combined-results}\footnote{For improved readability, some rows are duplicated when they correspond to the same run. LoRA Rank - Rank 64 is the same run as LoRA Modules - QKVO, which also corresponds to Interpolation - Quadratic.}.

We also briefly explore the effect of using a different VDM, running both LTX-Video \cite{hacohen2024ltx} and CogVideoX1.5 \cite{yang2024cogvideox}, see Table \ref{tab:ablations/vdm-combined}. This confirms our intuition that generalization and few-shot performance emerge and scale with model size, similar to LLMs \cite{brown2020language, llm_emergent}.

\begin{table}[t]
  \caption{Comparison of the two different VDMs across Binary Segmentation (mIoU ↑) and Pose Estimation (Match Rate ↑) over increasing number of training samples. \bestnt{Best} and \secondnt{second-best} results are highlighted.}
  \label{tab:ablations/vdm-combined}
  \centering
  \scriptsize
  \renewcommand{\arraystretch}{1.2}
  \setlength{\tabcolsep}{5pt}
  \begin{tabular}{lcccccccc}
    \toprule
    & \multicolumn{4}{c}{\textbf{Binary Segmentation (mIoU ↑)}} & \multicolumn{4}{c}{\textbf{Pose Estimation (Match Rate ↑)}} \\
    \cmidrule(lr){2-5} \cmidrule(lr){6-9}
    \textbf{VDM} & \(\mathbf{n=3}\) & \(\mathbf{n=5}\) & \(\mathbf{n=10}\) & \(\mathbf{n=30}\) & \(\mathbf{n=3}\) & \(\mathbf{n=5}\) & \(\mathbf{n=10}\) & \(\mathbf{n=30}\) \\
    \midrule
    LTX-Video (2B) & \second{16.41} & \second{20.78} & \second{35.39} & \second{46.15} & \second{14.64} & \second{21.04} & \second{26.41}  &  \second{38.47}   \\
    CogVideoX1.5 (5B) & \best{34.08} & \best{40.78} & \best{48.45} & \best{58.16} &\best{46.38} & \best{51.34} & \best{59.31} & \best{65.09}   \\
    % Cosmos & X & X & X & X & - & - & - & - \\
    \bottomrule
  \end{tabular}
\end{table}

\subsection{Visual Reasoning}\label{sec:visual_reasoning}

\subsubsection{ARC-AGI}

The ARC-AGI benchmark \cite{chollet2019measure} evaluates abstract reasoning in agents using tasks that demand compositional understanding, few-shot learning, and inductive reasoning. Each task comprises a small set of input-output image pairs (typically 2–5), from which the model must infer the underlying transformation rule to predict the correct output for novel inputs.

To adapt ARC to our setting, we reformulate each task as a video generation problem: both inputs and outputs are rendered as RGB images, and the model generates videos by interpolating between these images in a discrete way, \emph{i.e.}, with a sharp transition from the input to the output at half-way of the sequence (see \textit{discrete} in Figure \ref{fig:design-choices/interpolation-method-a}).

\begin{figure}[htbp]
  \centering
  \includegraphics[width=1.0\linewidth]{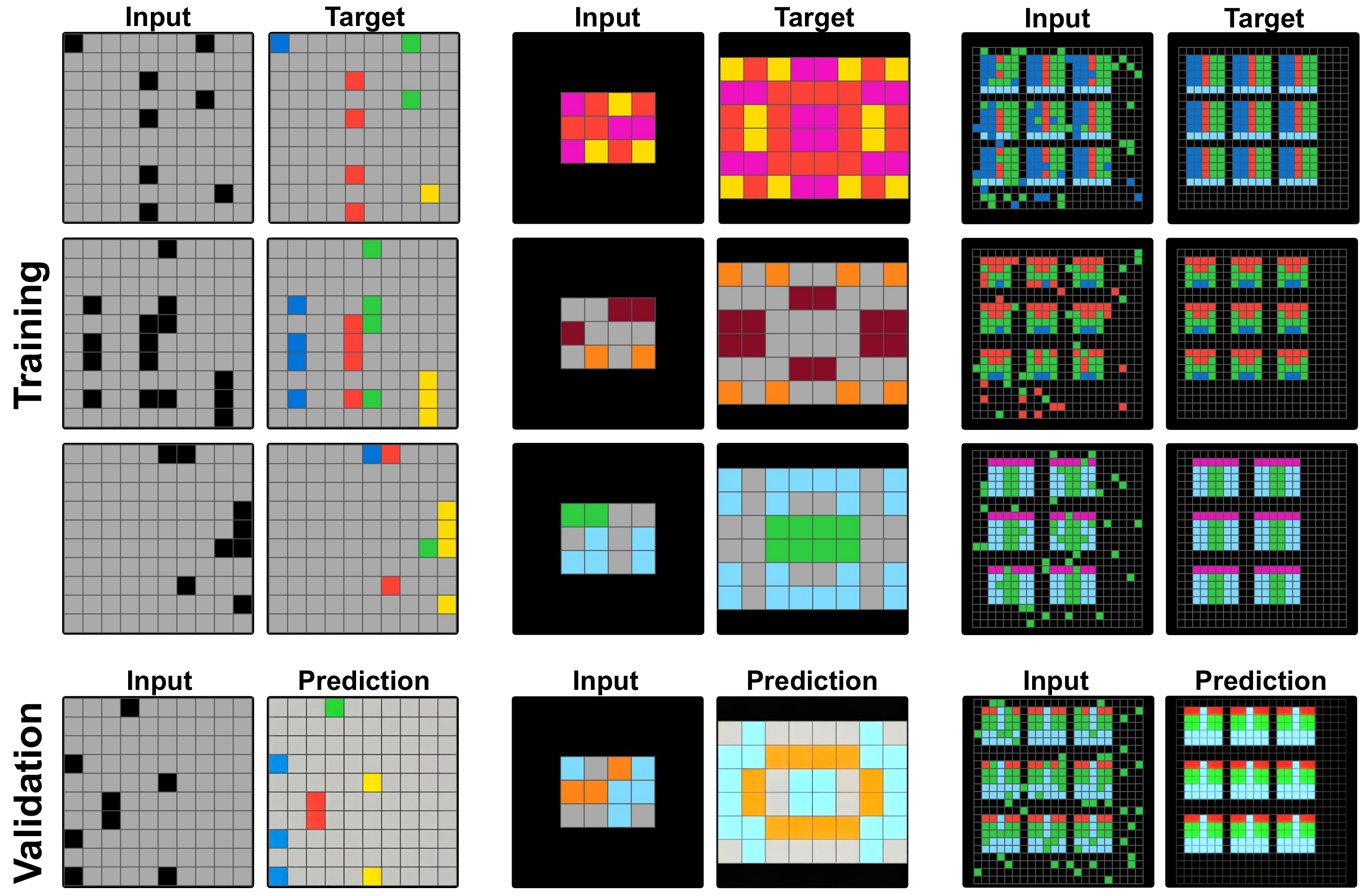}
  \caption{Examples of solved tasks from ARC. Training samples for each task (first three rows), followed by evaluation (last row).}
  \label{fig:arc-agi/qualitative-examples}
\end{figure}

\begin{figure}[htbp]
  \centering
  \begin{minipage}{0.65\linewidth}
    \centering
    \includegraphics[width=\linewidth]{
    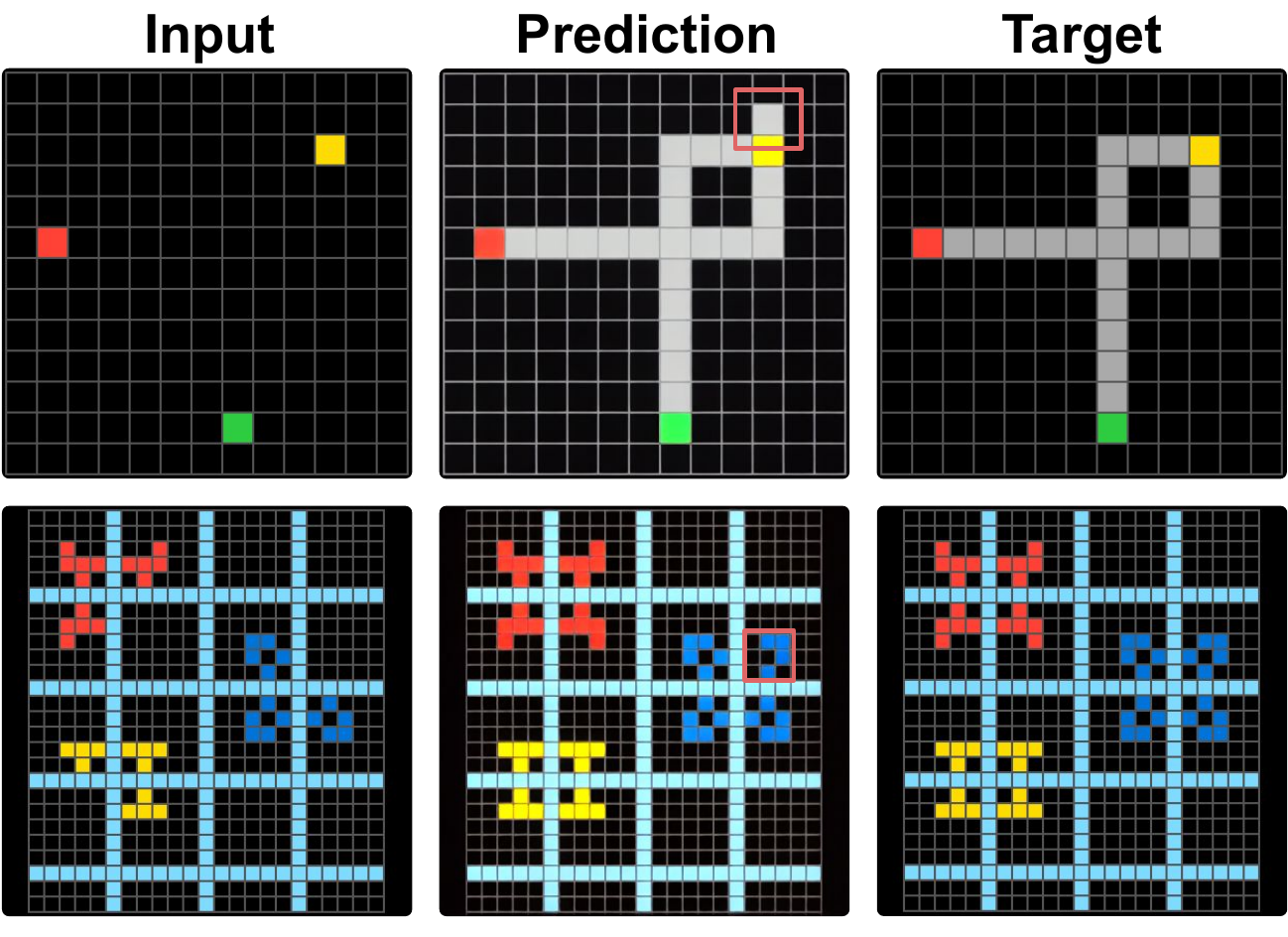}
    \subcaption{}
    \label{fig:arc-agi/rel_perf_evol}
  \end{minipage}%
  \hfill
  \begin{minipage}{0.33\linewidth}
    \centering
    \includegraphics[width=\linewidth]{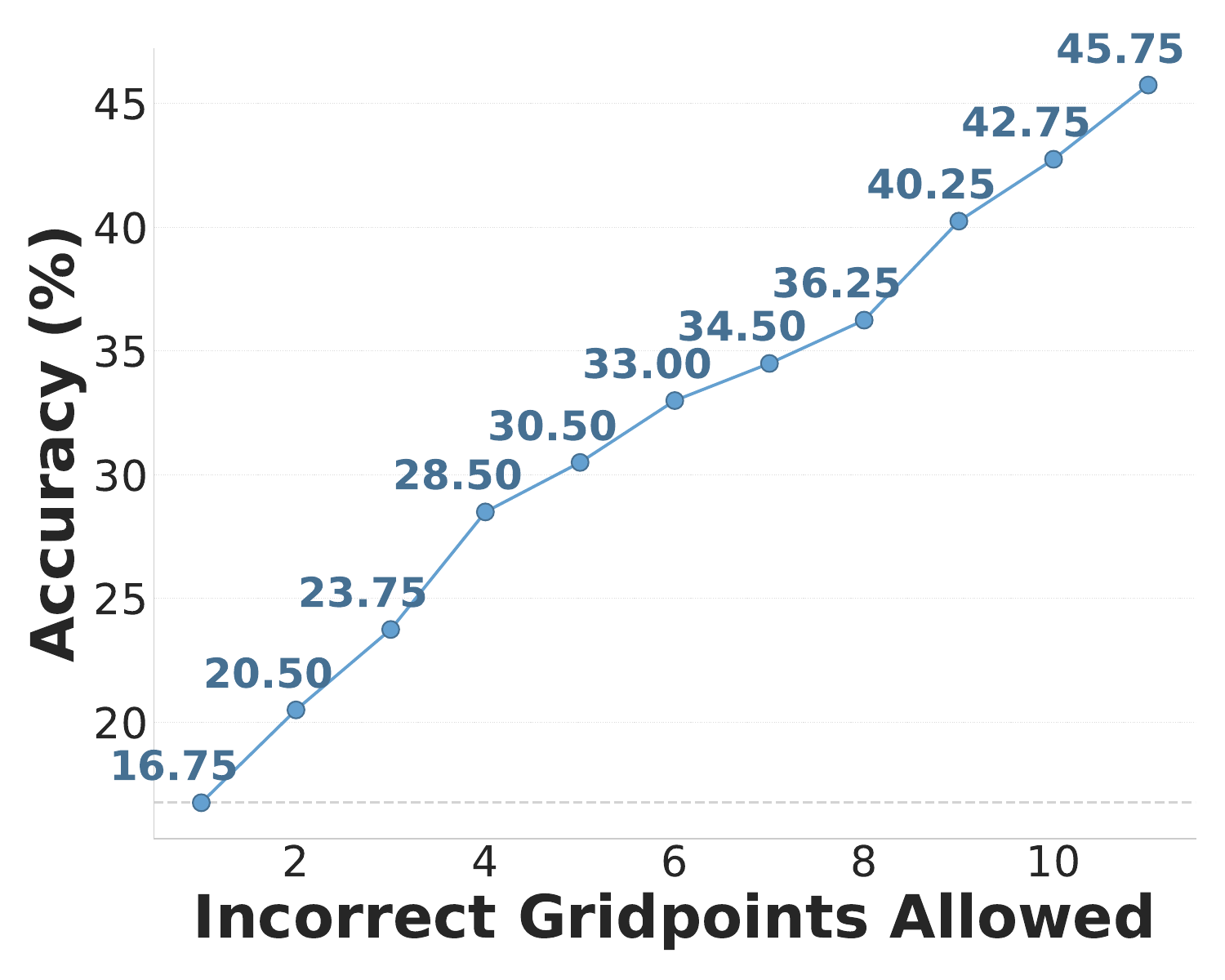}
    \subcaption{}
    \label{fig:arc-agi/arc_one_off}
    \vspace{0.5em}
    \includegraphics[width=\linewidth]{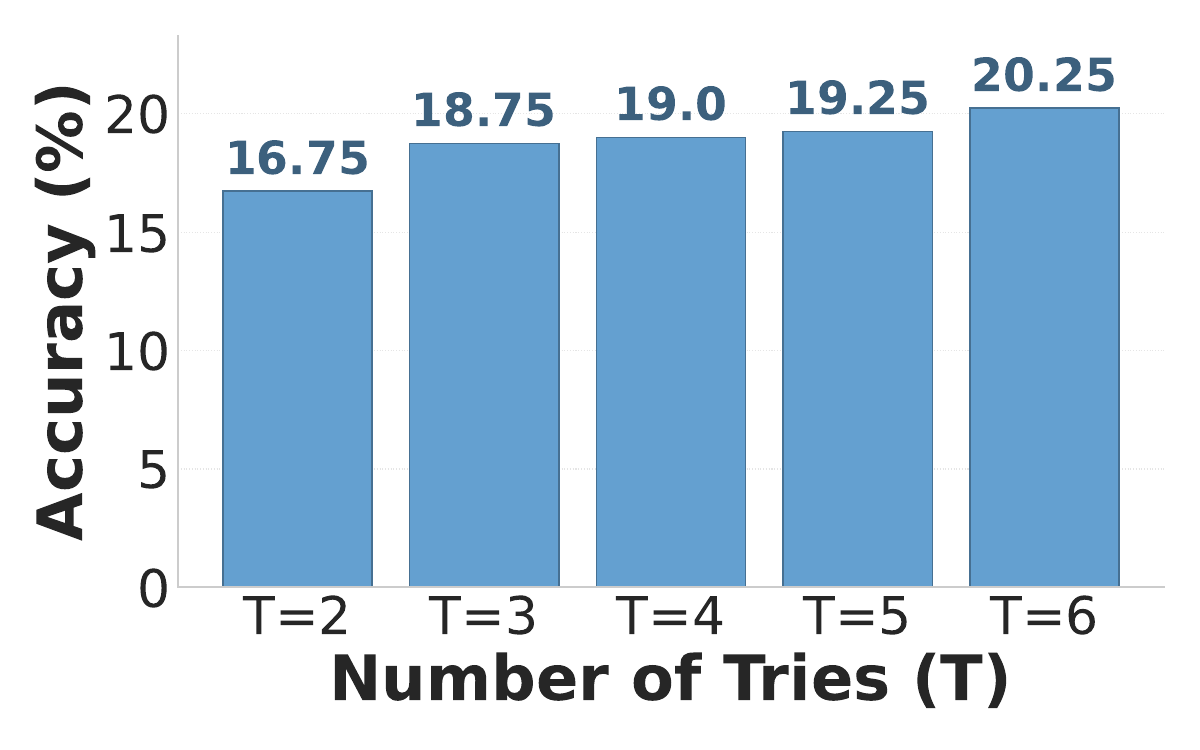}
    \subcaption{}
    \label{fig:arc-agi/perf_n_tries}
  \end{minipage}
  \caption{(a) Samples where the best model prediction failed on a single grid point, (b) Accuracy calculation allowing samples with incorrect grid points to be considered valid (2 attempts), and (c) Change in accuracy with increasing number of model attempts.}
  \label{fig:arc-agi/lenient-evaluation}
\end{figure}

For evaluation, we compare our method against direct prompting approaches using LLMs@pass1. We adopt the evaluation protocol from \cite{chollet2024arc}, which allows only two attempts per test input and accepts an answer only if it exactly matches the ground truth on all eval samples for that question. Quantitative results are presented in Table \ref{tab:arc-model-comparison}, with qualitative examples collected in Figure \ref{fig:arc-agi/qualitative-examples}. We also demonstrate how model performance changes when being progressively more lenient in the evaluation (see Figure~\ref{fig:arc-agi/lenient-evaluation}). 

While our method is far from achieving state-of-the-art performance on ARC \cite{akyurek2025surprisingeffectivenesstesttimetraining, berman2024record}, it demonstrates the viability of video-based reasoning as an emerging paradigm for visual reasoning tasks. Importantly, the model \textbf{has not been pre-trained on any data resembling ARC-AGI}, yet still manages to produce coherent, in-distribution outputs, 
even in the cases when it gives incorrect answers.

\begin{table}[htbp]
    \centering
    \caption{Accuracy on ARC public eval; CogVideoX1.5 fine-tuned with \colorbox{blue!10}{our method} highlighted.}
    \begin{tabular}{lc}
        \toprule
        \textbf{Model} & \textbf{Accuracy (\%)}\\
        \midrule
        OpenAI o1-preview & 21.00 \\
        Anthropic Claude 3.5 Sonnet & 21.00 \\
        \rowcolor{blue!10} CogVideoX1.5 & 16.75 \\
        OpenAI GPT-4o & \phantom{0}9.00 \\
        Google Gemini 1.5 & \phantom{0}8.00 \\
        \bottomrule
    \end{tabular}
    \label{tab:arc-model-comparison}
\end{table}

\subsubsection{ConceptARC}

\begin{figure}[htbp]
  \centering
  \includegraphics[width=0.8\linewidth]{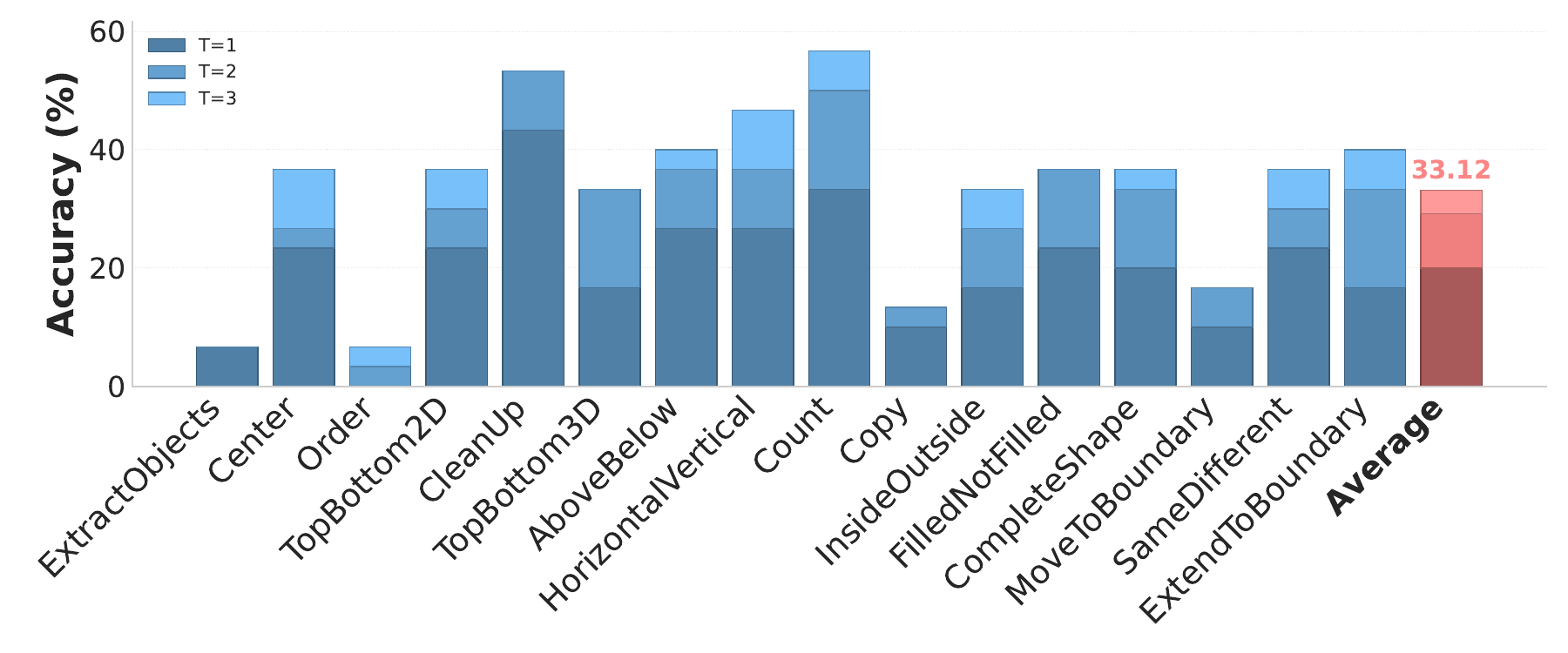}
  \caption{Accuracy in ConceptARC with different number of tries.}
  \label{fig:arc-agi/concept-arc-progression}
\end{figure}

To further explore visual reasoning capabilities, we evaluate on ConceptARC \cite{moskvichev2023conceptarc}, a simplified version of ARC specifically designed to assess concept understanding and generalization on different concepts. ConceptARC organizes problems into 16 concept groups (such as ``Above and Below,'' ``Center,'' ``Count,'' etc.), with each concept group containing 10 tasks and each task having 3 unique test inputs.
Following the evaluation methodology established by \cite{moskvichev2023conceptarc}, we allow three attempts per test input, and consider a test input correctly solved if any of these three attempts produces the correct answer. Our results are collected in Figure~\ref{fig:arc-agi/concept-arc-progression}.

\section{Limitations}\label{sec:limitations}

\noindent\textbf{Computational Complexity}
Fine-tuning VDMs remains computationally intensive despite using LoRAs. Future work could explore composable LoRAs \cite{huang2024lorahub, prabhakar2024lorasoups}, to reduce retraining costs while enhancing generalization. The iterative nature of diffusion sampling adds computational overhead. While tasks like segmentation and pose estimation often perform well with few, or even a single, sampling step \cite{xu2024matters}, performance on others, such as ARC-AGI, tends to degrade.\\ 
\noindent\textbf{RGB Representation Constraints} 
Our approach requires encoding all tasks into and from RGB space, which presents limitations for tasks requiring different modalities or high precision color representations, making evaluation less meaningful on these tasks. Color shifts observed in predictions (see Figures ~\ref{fig:arc-agi/qualitative-examples}~and~\ref{fig:misc/image-transforms}) illustrate this challenge for tasks which require high RGB precision. Notably, we observed consistent color shifts between training and test samples, suggesting the potential of learning deterministic corrective mappings from the training data.

\section{Conclusion}
We present a general and sample-efficient framework for few-shot adaptation of VDMs to a wide range of visual tasks. By reformulating task demonstrations as transition videos and fine-tuning lightweight LoRA adapters, our method enables VDMs to internalize and replicate complex visual transformations from just a handful of examples. Beyond conventional generative use cases, we demonstrated that VDMs possess latent visual understanding that can be unlocked through minimal supervision, achieving promising results on tasks including geometric transformations, style transfer, and abstract visual reasoning.

These findings support a broader view of VDMs as general-purpose visual learners, not merely generative engines. Our work opens up new directions for developing video-based foundation models that can flexibly adapt to novel tasks through efficient adaptation strategies. We believe that video generation, with its inherent structural richness, offers a powerful inductive bias for unifying perception, generation, and reasoning in visual AI.

\section*{Acknowledgements}

This work was supported as part of the Swiss AI Initiative by a grant from the Swiss National Supercomputing Centre (CSCS) under project ID a03 and a144 on Alps. Pablo Acuaviva, Aram Davtyan and Sebastian Stapf were supported by SNSF Grant 10001278.
Some of the calculations were performed on UBELIX (https://www.id.unibe.ch/hpc), the HPC cluster at the University of Bern.

% References
\bibliographystyle{plain}
\bibliography{references}

\include{appendix}

\end{document}

%% file: appendix.tex
\section*{Appendix}
\appendix

% \tableofcontents

\section{Experimental Details} 

In this section, we provide comprehensive details regarding our experimental setup, methodology, and implementation to facilitate reproducibility of our results.

\subsection{Datasets}

\paragraph{COCO 2017 \cite{lin2014microsoft}} 
We use the COCO 2017 dataset for binary segmentation and human pose estimation. The dataset provides richly annotated images with object masks and keypoints, allowing for diverse vision task evaluation. We employ a simple script to convert binary masks into black and white images, and to color the different pose components (white head, blue torso, red arms and green legs) to generate the target images.

\paragraph{DreamBooth \cite{ruiz2023dreambooth}} 
We use DreamBooth to evaluate models on a range of vision tasks including geometric transformations and style transfer. We apply different transforms to prepare the images for training (convert to gray for colorization, remove squares for inpainting, etc).

\paragraph{TinyImageNet \cite{tiny-imagenet}} 
We use TinyImageNet for constructing the classification task grid. We select a subset of 7 categories with 100 images each at a resolution of 64x64. We select 80 images from each category for building the training grids and the remaining 20 for building the validation grids. We build 4x4 grids by randomly sampling images for the respective split.

\paragraph{ARC-AGI \cite{chollet2024arc}} 
We utilize the public evaluation subset of the ARC-AGI dataset, provided in JSON format with paired input-output examples for training and testing. A custom script converts the data to images and back, enabling visual processing and analysis.

\paragraph{ConceptARC \cite{moskvichev2023conceptarc}} 
Similarly to ARC-AGI, ConceptARC provides structured reasoning tasks in a visual format. We apply the same conversion and evaluation pipeline as with ARC-AGI.

\paragraph{Additional Datasets} In the appendix, we provide some additional results, for which we use the NYUv2 \cite{Silberman:ECCV12}, ADE20K \cite{zhou2017scene, zhou2019semantic}, and ML-Hypersim \cite{roberts:2021} datasets.

\subsection{Hyperparameters}

We list the main hyperparameters used in our experiments in Table~\ref{appdx:tab:hyperparameters}. These hyperparameters remain consistent across all experiments unless explicitly stated otherwise (e.g., in ablation studies). When a configuration differs for a specific task, it is indicated with a dash (–), and further details are provided in the corresponding tables (see Tables~\ref{appdx:tab:cogvideo_training_details} and \ref{appdx:tab:cogvideo_resolutions}).

\begin{table}[ht]
\centering
\caption{Comparison of hyperparameters used in CogVideoX1.5 and LTX experiments}
\label{appdx:tab:hyperparameters}
\begin{tabular}{lll}
\toprule
\textbf{Parameter} & \textbf{CogVideoX1.5} & \textbf{LTX} \\
\midrule
\multicolumn{3}{l}{\textit{LoRA Configuration}} \\
Rank & 64 & 64 \\
Alpha & 32 & 32 \\
Target modules & QKVO & All linear \\
\midrule
\multicolumn{3}{l}{\textit{Video Configuration}} \\
Resolution (FxHxW) & - & 9x480x704 \\
Interpolation & - & Convex \\
\midrule
\multicolumn{3}{l}{\textit{Training Configuration}} \\
Seed & 42 & 42 \\
Batch size & 2 & 16 \\
Training steps & – & 3000 \\
Validation steps & – & 200 \\
Gradient accumulation steps & 1 & 1 \\
\midrule
\multicolumn{3}{l}{\textit{Optimizer Configuration}} \\
Optimizer & AdamW & AdamW \\
Learning rate & 1e-4 & 1e-3 \\
Scheduler & Constant & Constant \\
\(\beta_1\) & 0.90 & 0.90 \\
\(\beta_2\) & 0.95 & 0.99 \\
Weight decay & 1e-3 & 0.0 \\
\(\epsilon\) & 1e-8 & 1e-8 \\
Max grad norm & 1.0 & 1.0 \\

\bottomrule
\end{tabular}
\end{table}

\begin{table}[ht]
\centering
\caption{Steps details for CogVideoX1.5 across different tasks.}
\label{appdx:tab:cogvideo_training_details}
\begin{tabular}{lll}
\toprule
\textbf{Task} & \textbf{Training Steps} & \textbf{Validation Steps} \\
\midrule
\multicolumn{3}{l}{\textbf{Binary Segmentation}} \\
\(n=3\)  & 1000 &  1000  \\
\(n=5\)  & 1000 &  1000  \\
\(n=10\)  & 2000 &  1000  \\
\(n=30\) & 4000 &  1000  \\
\midrule
\multicolumn{3}{l}{\textbf{Pose Estimation}} \\
\(n=3\)  & 1000 &  1000  \\
\(n=5\)  & 1000 &  1000  \\
\(n=10\)  & 2000 &  1000  \\
\(n=30\) & 4000 &  1000  \\
\midrule
ARC AGI & 1800 & 300  \\
ConceptARC & 600 &  100  \\
Classification TinyImageNet & 500 &  500  \\
Other tasks & 1200 & 300  \\
\bottomrule
\end{tabular}
\end{table}

\begin{table}[ht]
\centering
\caption{Resolution configurations for CogVideoX1.5 across tasks.}
\label{appdx:tab:cogvideo_resolutions}
\begin{tabular}{ll}
\toprule
\textbf{Task} & \textbf{Resolution (FxHxW)} \\
\midrule
Binary Segmentation & 9x480x720 \\ 
Pose Estimation & 9x480x720 \\
ARC AGI & 9x480x480 \\
ConceptARC & 9x480x480 \\
Classification TinyImageNet & 9x256x256 \\
Other tasks & 9x480x720 / 9x480x480  \\
\bottomrule
\end{tabular}
\end{table}

Beyond the hyperparameters detailed in Table~\ref{appdx:tab:hyperparameters}, additional inference-time adjustments were made to improve computational efficiency. Notably, for experiments involving CogVideoX1.5, Classifier-Free Guidance (CFG) was not used during inference. Moreover, for ablation studies with CogVideoX1.5 on Binary Segmentation and Pose Estimation, the number of diffusion sampling steps was decreased from a default of 50 to 5, a change made after confirming that it yielded comparable results following fine-tuning.

\subsection{Computational Requirements} All experiments were conducted using NVIDIA RTX 4090 GPUs with 24GB of VRAM. We utilized a multi-GPU compute node to execute multiple tasks in parallel, though each individual run was performed on a single GPU. Table~\ref{appdx:tab:computational_requirements} details the computational resources required for general tasks. For ablation studies on Binary Segmentation and Pose Estimation, we report compute times in Table~\ref{appdx:tab:computational_requirements_ablation}. Comparisons between CogVideoX1.5 and LTX-Video computational times are collected in Table \ref{appdx:tab:computational_requirements_model_comparison}. 

\begin{table}[ht]
\centering
\caption{Training Time (in hours) for CogVideoX1.5 on general tasks}
\label{appdx:tab:computational_requirements}
\begin{tabular}{l c}
\toprule
\textbf{Task} & \textbf{Total (h)} \\
\midrule
ARC AGI        & 450.00 \\
Concept ARC    & 130.00 \\
TinyImageNet   &  25.00 \\
Other Tasks    & 150.00 \\
\bottomrule
\end{tabular}
\end{table}

\begin{table}[ht]
\centering
\caption{
Training times for CogVideoX1.5 ablations across number of training samples. Total reflects added time for all ablations.
}
\label{appdx:tab:computational_requirements_ablation}
\begin{tabular}{lccc}
\toprule
\textbf{Task} & \textbf{Training Time (h)} & \textbf{\# Runs} & \textbf{All runs (h)} \\
\midrule
\multicolumn{4}{l}{\textbf{Binary Segmentation}} \\
\(n=3\)  & 0.75 & 3 & 2.25 \\
\(n=5\)  & 0.75 & 3 & 2.25 \\
\(n=10\) & 1.50 & 3 & 4.50 \\
\(n=30\) & 3.00 & 1 & 3.00 \\
\midrule
\multicolumn{4}{l}{\textbf{Pose Estimation}} \\
\(n=3\)  & 1.25 & 3 & 3.75 \\
\(n=5\)  & 1.25 & 3 & 3.75 \\
\(n=10\) & 2.50 & 3 & 7.50 \\
\(n=30\) & 5.00 & 1 & 5.00 \\
\midrule
\multicolumn{3}{l}{\textbf{Time Per Ablation}} & 32.00 \\
\multicolumn{3}{l}{\textbf{Total}} & 352.00 \\
\bottomrule
\end{tabular}
\end{table}

\begin{table}[ht]
\centering
\caption{
Training times for CogVideoX1.5 and LTX-Video across tasks and number of training samples. Total time reflects cumulative runtime for each configuration.
}
\label{appdx:tab:computational_requirements_model_comparison}
\begin{tabular}{lcccccc}
\toprule
\textbf{Task} 
& \multicolumn{3}{c}{\textbf{CogVideoX1.5}} 
& \multicolumn{3}{c}{\textbf{LTX-Video}} \\
\cline{2-7}
 & Training Time (h) & \# Runs & All runs (h) 
 & Training Time (h) & \# Runs & All runs (h) \\
\midrule
\multicolumn{7}{l}{\textbf{Binary Segmentation}} \\
\(n=3\)  & 0.50  & 3 & 1.50 & 1.5 & 3 & 4.50\\
\(n=5\)  & 0.50 &  3&  1.50&  1.5 & 3 & 4.50\\
\(n=10\) & 1.00 &  3&  3.00&  1.5 & 3 & 4.50 \\
\(n=30\) & 2.00 &  1 & 2.00 & 1.5 & 1 & 1.50 \\
\midrule
\multicolumn{7}{l}{\textbf{Pose Estimation}}   \\
\(n=3\)  & 0.75  & 3 & 2.25 &  1.75 & 3 & 5.25 \\
\(n=5\)  & 0.75 &  3&  2.25 &  1.75 & 3 & 5.25 \\
\(n=10\) & 1.50 &  3&  4.50 &  1.75 & 3 & 5.25 \\
\(n=30\) & 3.00 &  1 & 3.00 &  1.75 & 1 & 1.75 \\\midrule
\multicolumn{1}{l}{\textbf{Total}}
&  &  &  20.00 & &  & 32.50 \\
\bottomrule
\end{tabular}
\end{table}

\paragraph{Development Compute Time.} Approximately 100 GPU hours were additionally spent on tuning, debugging, and system integration across all tasks. These were not included in the tables above, which report only the finalized experimental runs.

\subsection{Evaluation}

We evaluate all tasks using standard metrics, with appropriate references in the main paper where applicable. The only exception is pose estimation, for which we introduce a new metric, the \textbf{Match Rate}. Below, we motivate this choice and detail its computation. For selected tasks, we also include additional evaluation details where they enhance clarity or reproducibility.

\subsubsection{Match Rate}

Our primary reason for deviating from standard pose estimation metrics is computational complexity and feasibility. Converting an RGB prediction into a format suitable for conventional evaluation involves significant overhead. Two common approaches are:

\begin{enumerate}
    \item \textit{Color-based encoding:} Assigning a unique color to each keypoint and decoding them from the RGB image. This is impractical, as it requires 17 visually distinct colors. Given that our fine-tuning process may induce color shifts, maintaining this separation is unreliable. Moreover, this method does not easily recover the grouping of keypoints into individual persons without additional post-processing heuristics.
    
    \item \textit{Per-keypoint prediction:} Generating a separate output for each keypoint and aggregating them afterward. However, this increases inference costs by nearly a factor of 20, which is prohibitive for our setup where inference speed is already a bottleneck.
\end{enumerate}

Importantly, our goal is not to compete with state-of-the-art pose estimation methods, but to demonstrate that the model possesses this capability. To this end, we introduce a computationally efficient metric, \textbf{Match Rate}, which is well correlated with pose estimation quality.

The metric is computed as follows:

\begin{enumerate}
    \item Given a predicted pose, we use the color channels to segment the figure into four main components: head, torso, arms, and legs.
    \item For each connected component, we compute its centroid.
    \item We determine the best possible matching between predicted centroids and the corresponding centroids derived from annotated data for each body part.
    \item A match is considered valid if the Euclidean distance between the predicted and annotated centroids is below a fixed multiple (1.5 in our experiments) of the average inter-head distance, computed from annotated keypoints. If no heads are available, we use a default threshold of 20 pixels for average inter-head distance.
\end{enumerate}

\begin{figure}[t]
  \centering
  \includegraphics[width=1.0\linewidth]{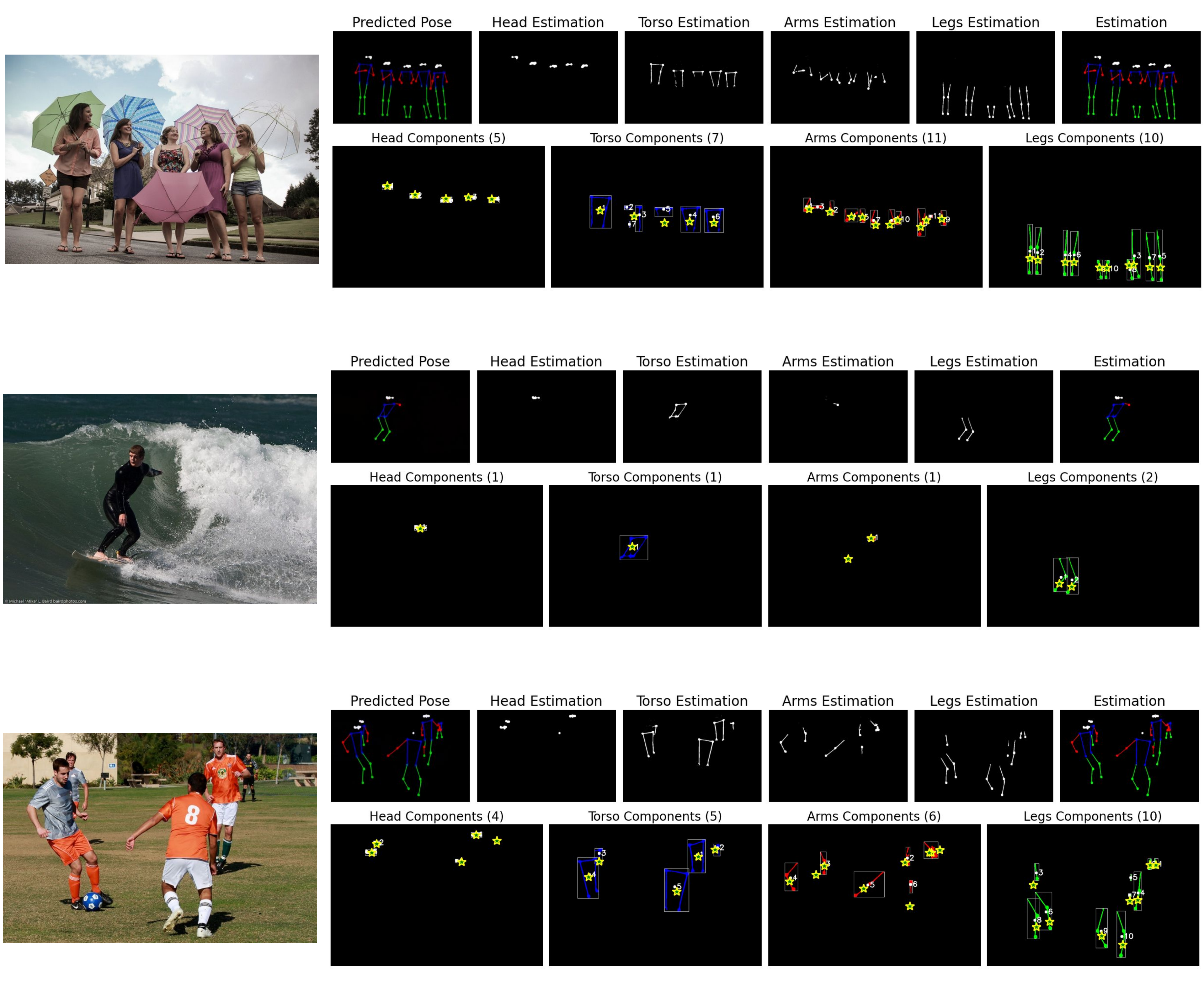}
  \caption{Match Rate calculation. Left: original image. Top row: model prediction and estimated components. Bottom row: centroids from estimations; annotated centroids shown as stars.}
  \label{appdx:fig:match_rate_calculation_0}
\end{figure}

We illustrate this process in Figure~\ref{appdx:fig:match_rate_calculation_0}. The left panel shows the original image, while the top row shows the predicted components. The bottom row displays, for each component, the predicted centroids (white numbered dots) and the ground-truth centroids from annotations (stars).

While effective, the Match Rate metric has some limitations. First, it assumes that the image is rendered over a black background to enable reliable component decoding. In our experiments, this assumption holds consistently, making the metric robust in practice. Second, the metric only captures matches, meaning that excessive or spurious predictions are not penalized. Although one might address this by penalizing unmatched predictions, decoding from RGB can occasionally yield multiple components for a single body part (e.g., the torso in the first image of Figure~\ref{appdx:fig:match_rate_calculation_0}), making such penalization potentially unfair. As we have visually verified that the model rarely produces false positives outside the intended figure, we do not believe this limitation meaningfully impacts the metric's reliability.

The main advantage of the Match Rate is that it allows us to evaluate pose estimation using only a single RGB prediction, offering a simple yet effective proxy for pose quality.

\subsubsection{ARC-AGI and ConceptARC}

To facilitate evaluation, we use a predefined number of training steps and conducted evaluations at regular intervals. From these evaluation checkpoints, our process proceeded as follows:

At each evaluation step, if the model's prediction on the training set was correct, we use the corresponding prediction and the appropriate number of subsequent steps as validation predictions: two for ARC-AGI and three for ConceptARC. If no correct training prediction was observed, we select the final available predictions as the best guesses for validation. This approach is not adaptive to the difficulty of individual problems. For simpler problems, it may include redundant evaluations, whereas more difficult problems may benefit from additional training. Such limitations are occasionally observed in ConceptARC, although only in a few cases. Overall, we found that our fixed schedule of training and evaluation steps represents a reasonable compromise. It allows us to demonstrate the model's reasoning capabilities without over-optimizing. We believe that improvements in both performance and computational efficiency are possible through a refined evaluation strategy.

Decoding introduces further challenges in both ARC-AGI and ConceptARC. In some instances, the model outputs single-colored black grids of varying sizes. These are difficult to decode reliably from RGB space. In ConceptARC, we addressed this issue by converting such outputs to gray gridpoints when doing so did not affect the underlying puzzle logic. In ARC-AGI, we encountered only one such case. Visual inspection confirmed that the model's prediction was correct, and no modification was made.

We expect that only a small number of such cases remain unmodified across both datasets. These residual instances do not affect the core reasoning results demonstrated by our experiments.

A final limitation of our decoding procedure is the assumption that the output grid size is known in advance. In practice, this is a quantity the model must predict. We have visually confirmed that the model generally outputs the correct grid size in successful cases. It does not produce amplified or padded versions of the target grid with the correct solution embedded inside.

\subsubsection{TinyImageNet}

For each image, we generate a 4×4 grid of classification patches, resulting in 16 classification examples per image. To decode these, we apply L2 nearest neighbor matching against the set of known symbols. After obtaining predictions from all evaluation patches, we aggregate them and proceed as in a standard classification task.

\section{Additional Results}

In this section, we provide further qualitative and quantitative results for the tasks discussed in the main paper.

\begin{figure}[htbp]
  \centering
  \includegraphics[width=1.0\linewidth]{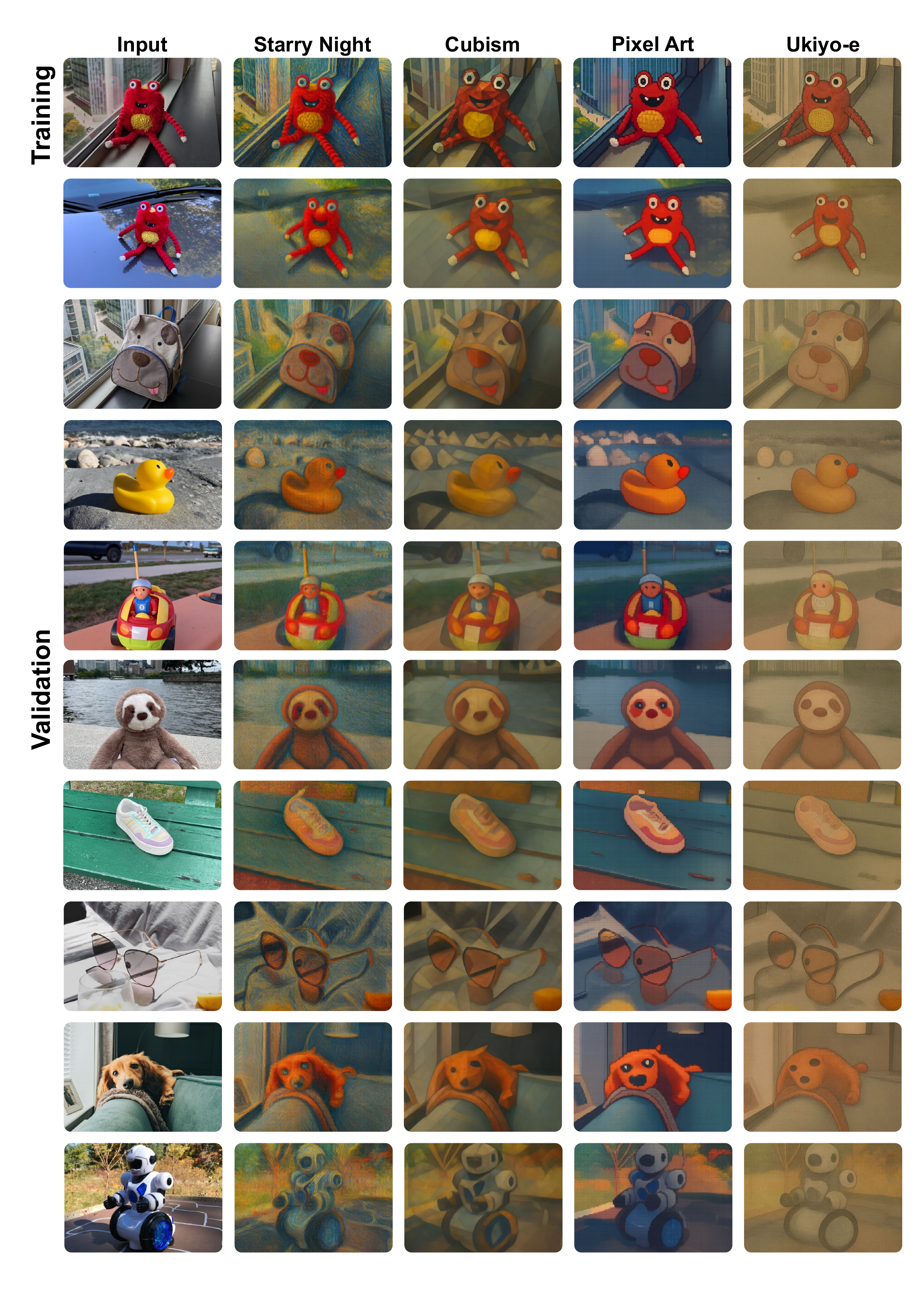}
  \caption{Additional qualitative results for Style Transfer. Training sample on first row, generated samples in subsequent rows.}
  \label{appdx:fig:misc-style-transfer}
\end{figure}

\begin{figure}[htbp]
  \centering
  \includegraphics[width=\linewidth]{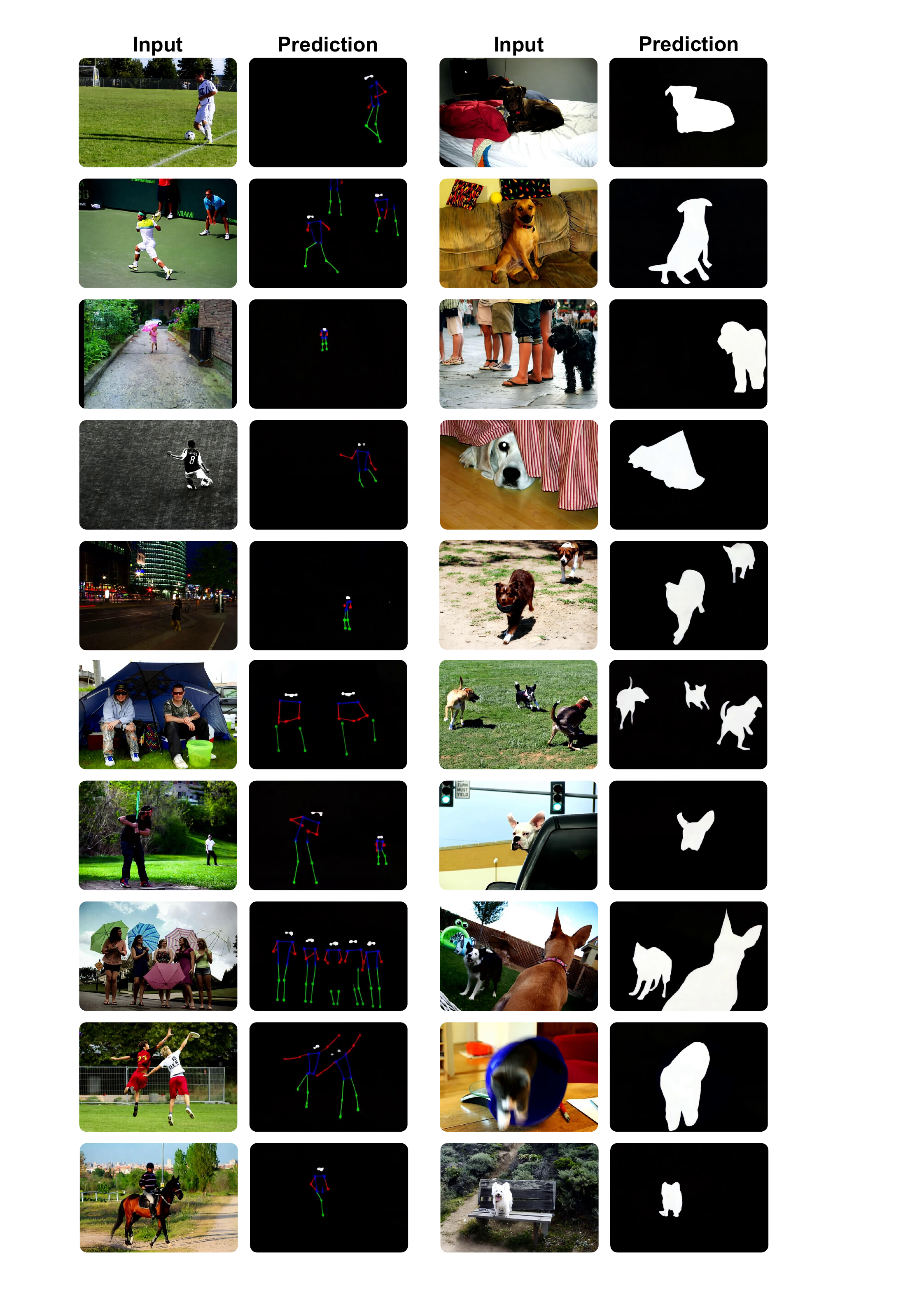}
  \caption{Predictions for pose estimation (left) and binary segmentation (right) for training with \(n = 30\) samples.}
  \label{appdx:fig:pose-binary-seg}
\end{figure}

Figure~\ref{appdx:fig:misc-style-transfer} shows additional samples for the style transfer task, illustrating how the model generalizes style across diverse content. Figure~\ref{appdx:fig:pose-binary-seg} presents outputs for pose estimation and binary segmentation when trained with \(n = 30\) samples.

\begin{figure}[htbp]
  \centering
  \includegraphics[width=\linewidth]{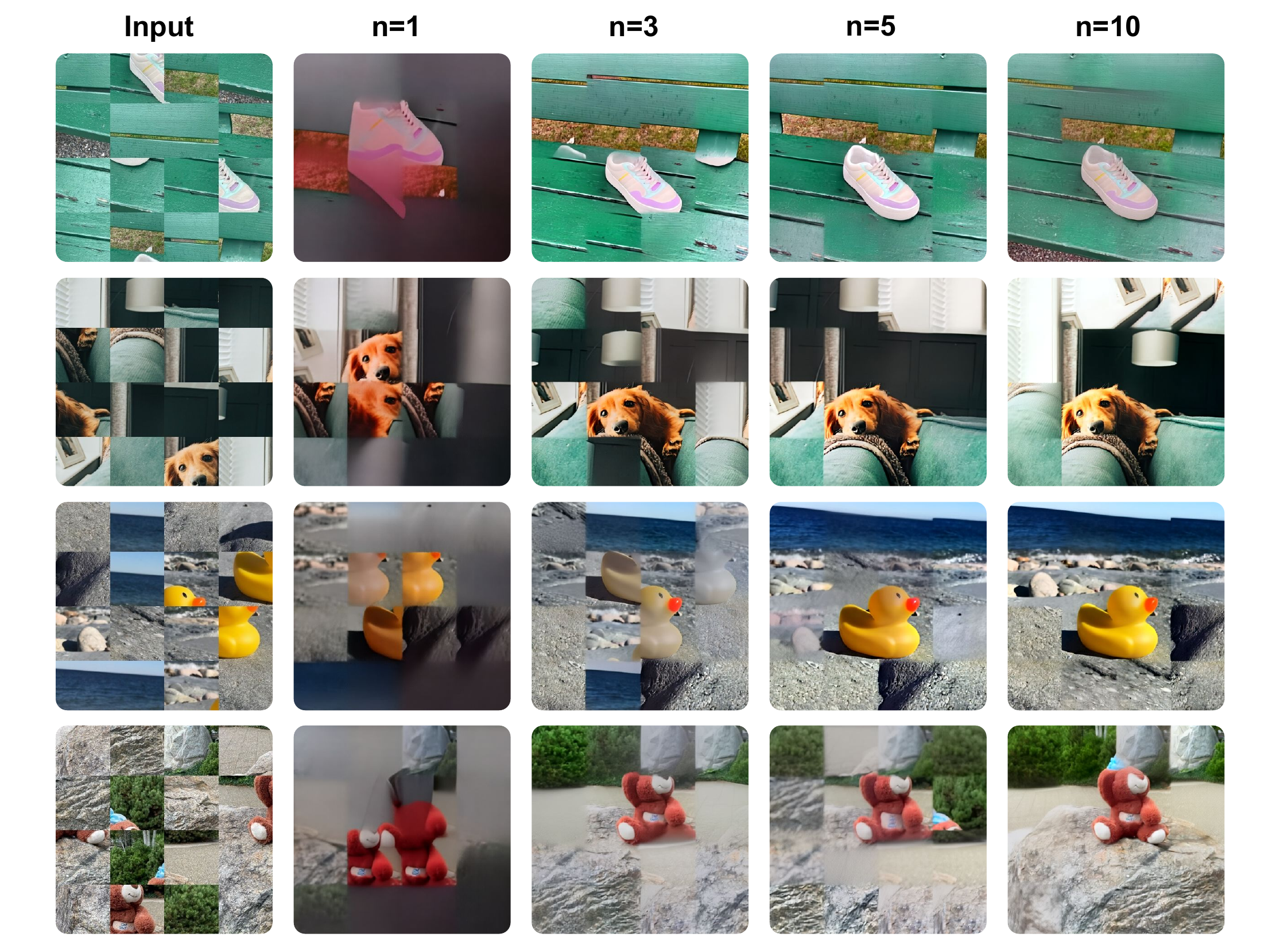}
  \caption{Progression of results on the Jigsaw task under varying numbers of training samples.}
  \label{appdx:fig:jigsaw}
\end{figure}

\begin{figure}[htbp]
  \centering
  \includegraphics[width=\linewidth]{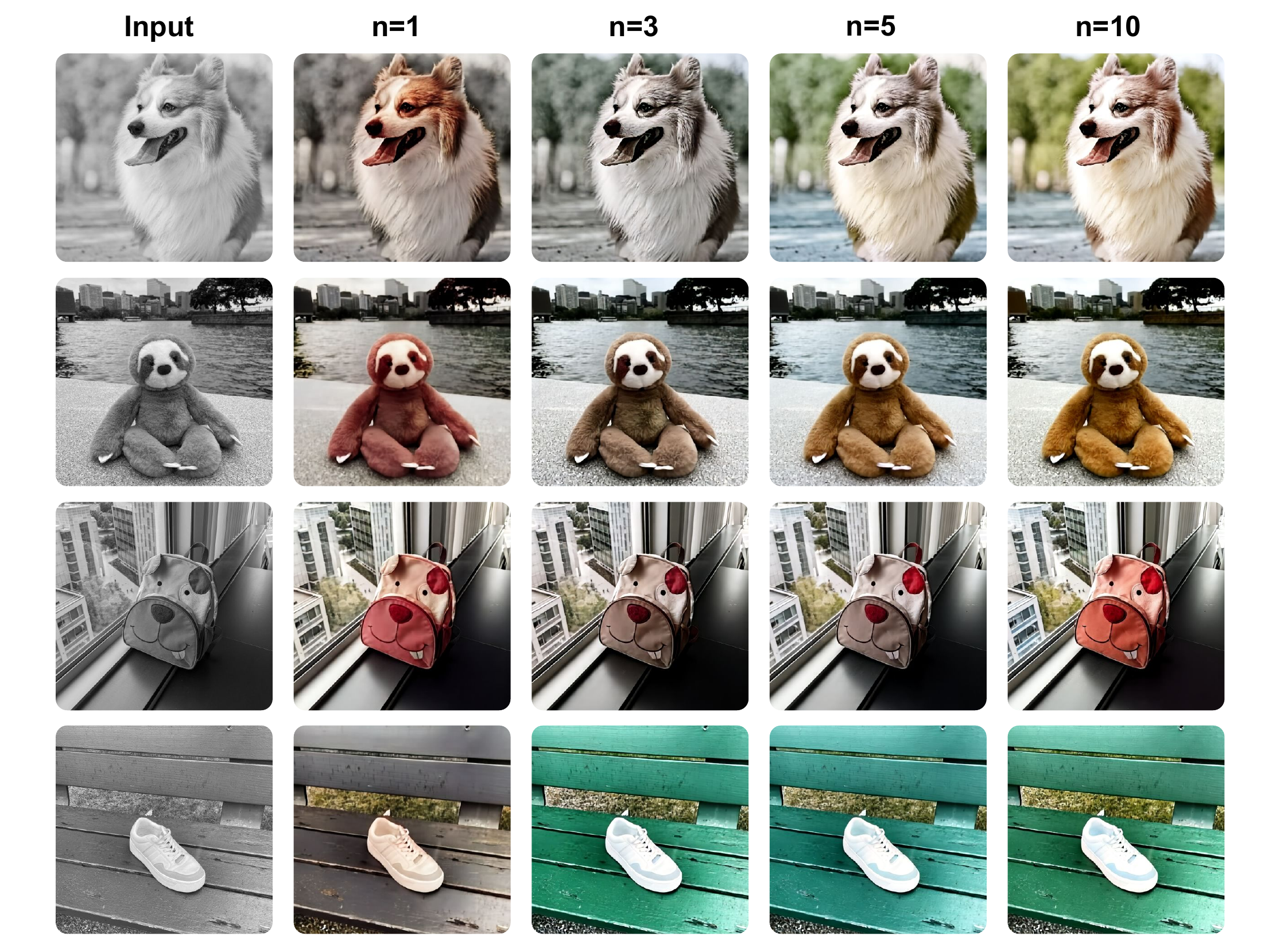}
  \caption{Progression of results on the Colorization task under varying numbers of training samples.}
  \label{appdx:fig:colorization}
\end{figure}

\begin{figure}[htbp]
  \centering
  \includegraphics[width=\linewidth]{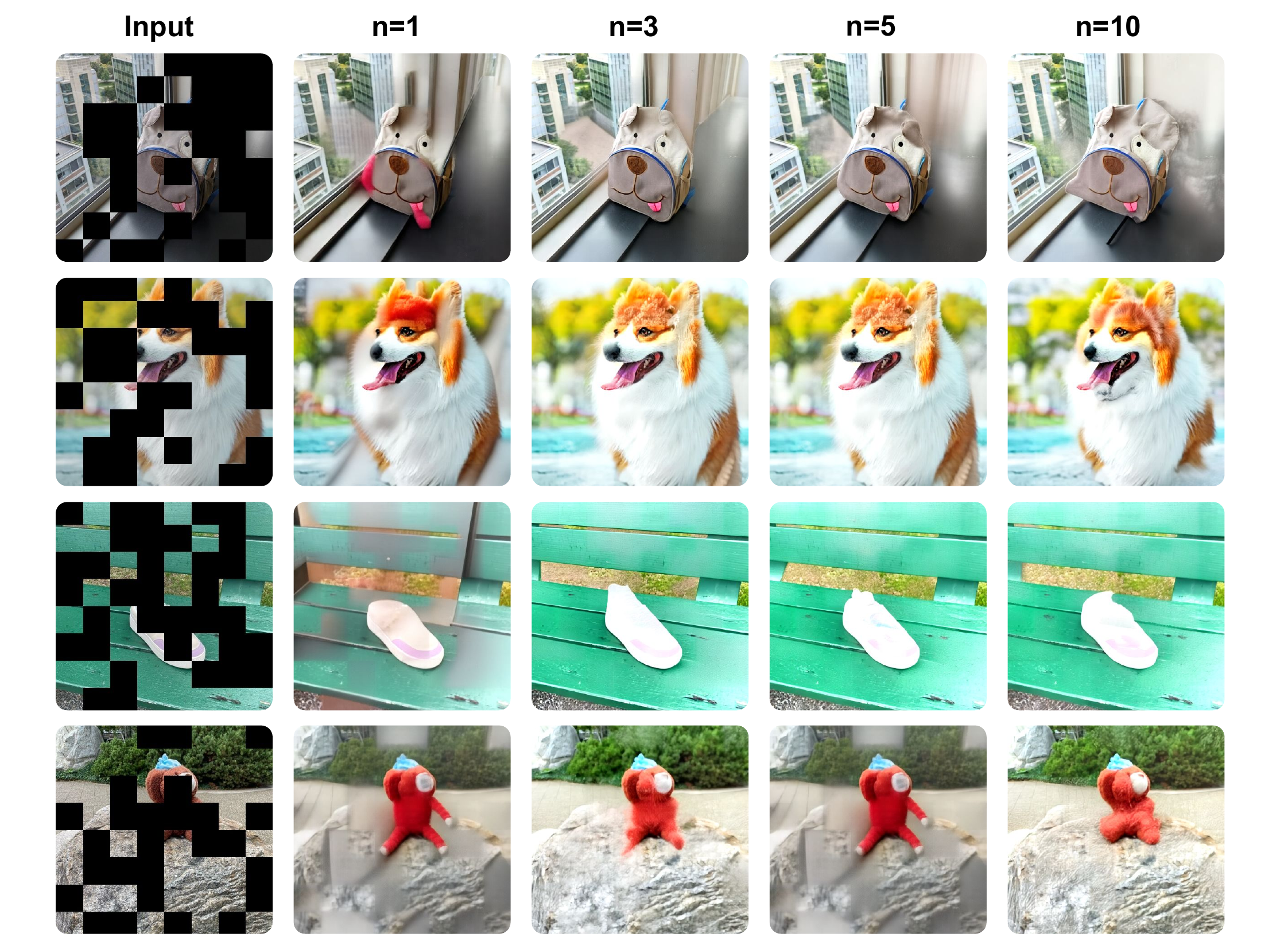}
  \caption{Progression of results on the Inpainting task under varying numbers of training samples.}
  \label{appdx:fig:inpainting}
\end{figure}

We show training progression on structured prediction tasks in Figures~\ref{appdx:fig:jigsaw},~\ref{appdx:fig:colorization}, and ~\ref{appdx:fig:inpainting}, including jigsaw puzzle solving, colorization, and inpainting. These highlight the improvements as the number of training samples increases.

\begin{figure}[htbp]
  \centering
  \includegraphics[width=\linewidth]{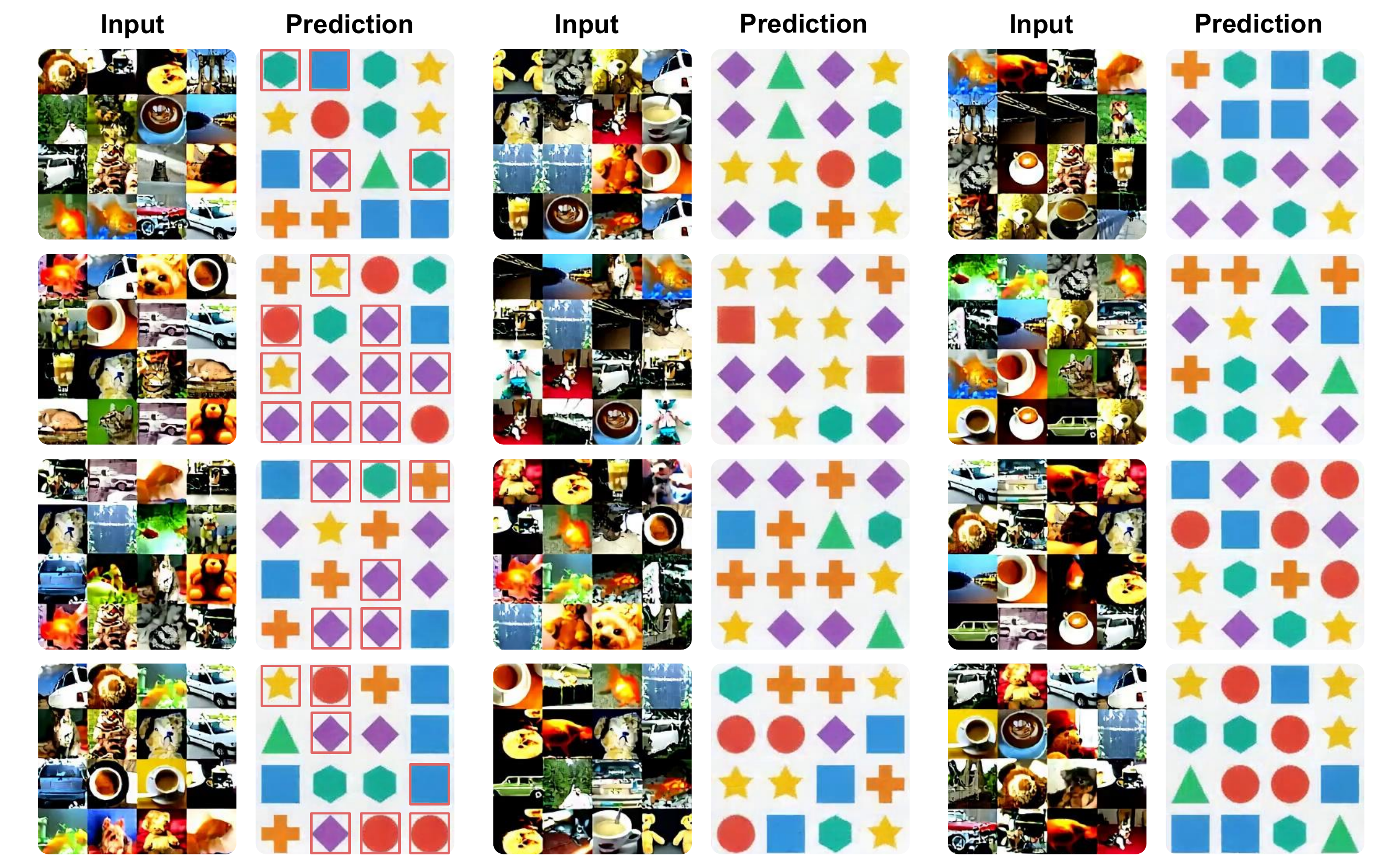}
  \caption{Validation results on the TinyImageNet grid classification task with \(n=30\) training samples. Prediction errors in the first column are highlighted with red boxes.}
  \label{appdx:fig:tiny-imagenet}
\end{figure}

In Figure~\ref{appdx:fig:tiny-imagenet} we show validation examples from the TinyImageNet classification task, where prediction errors are highlighted for interpretability.

\begin{figure}[htbp]
  \centering
  \includegraphics[width=0.8\linewidth]{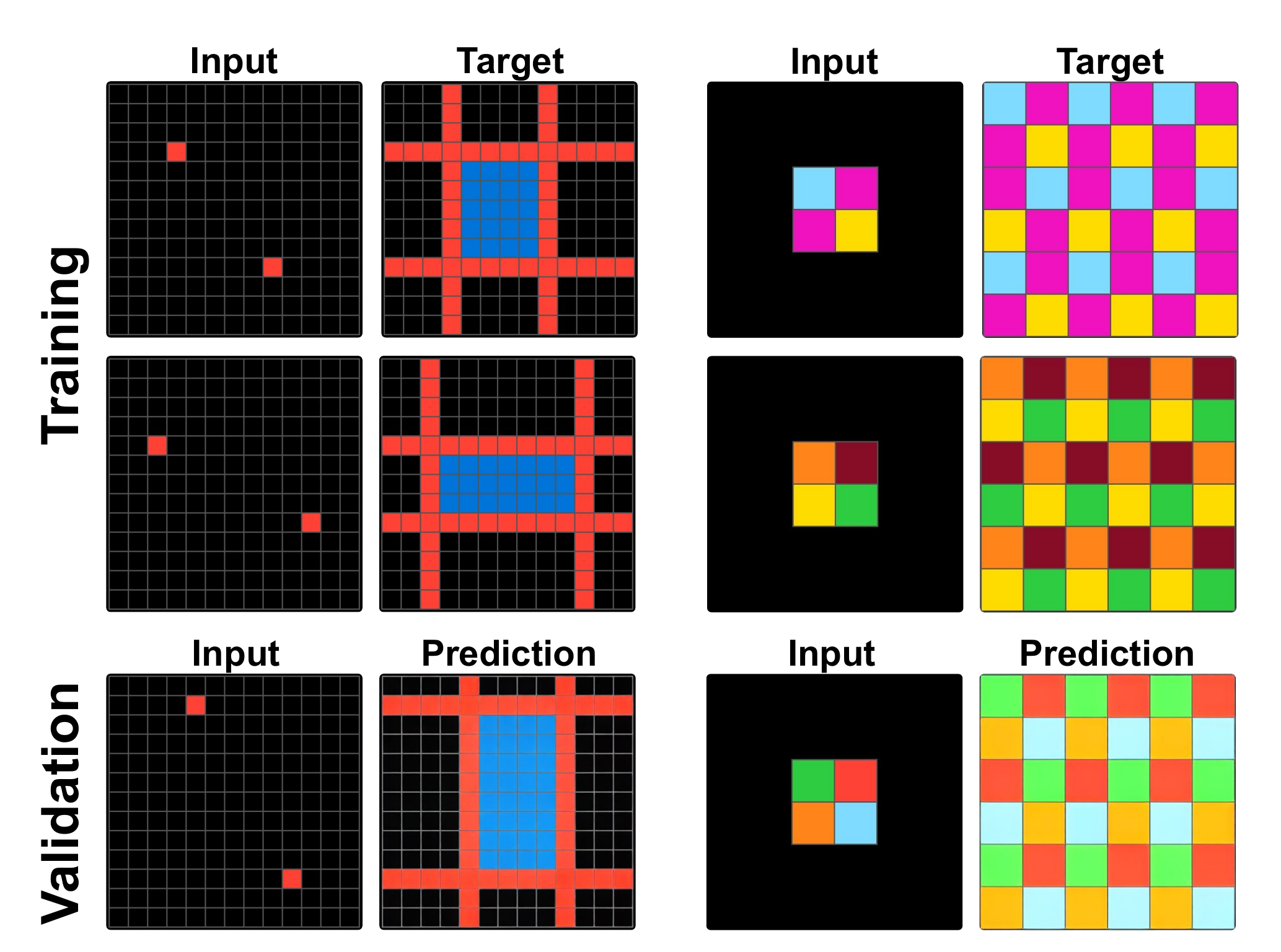}
  \caption{Additional examples of solved tasks from ARC-AGI. Training samples for each task (first two rows), followed by evaluation (last row).}
  \label{appdx:fig:arc-0}
\end{figure}

\begin{figure}[htbp]
  \centering
  \includegraphics[width=1.0\linewidth]{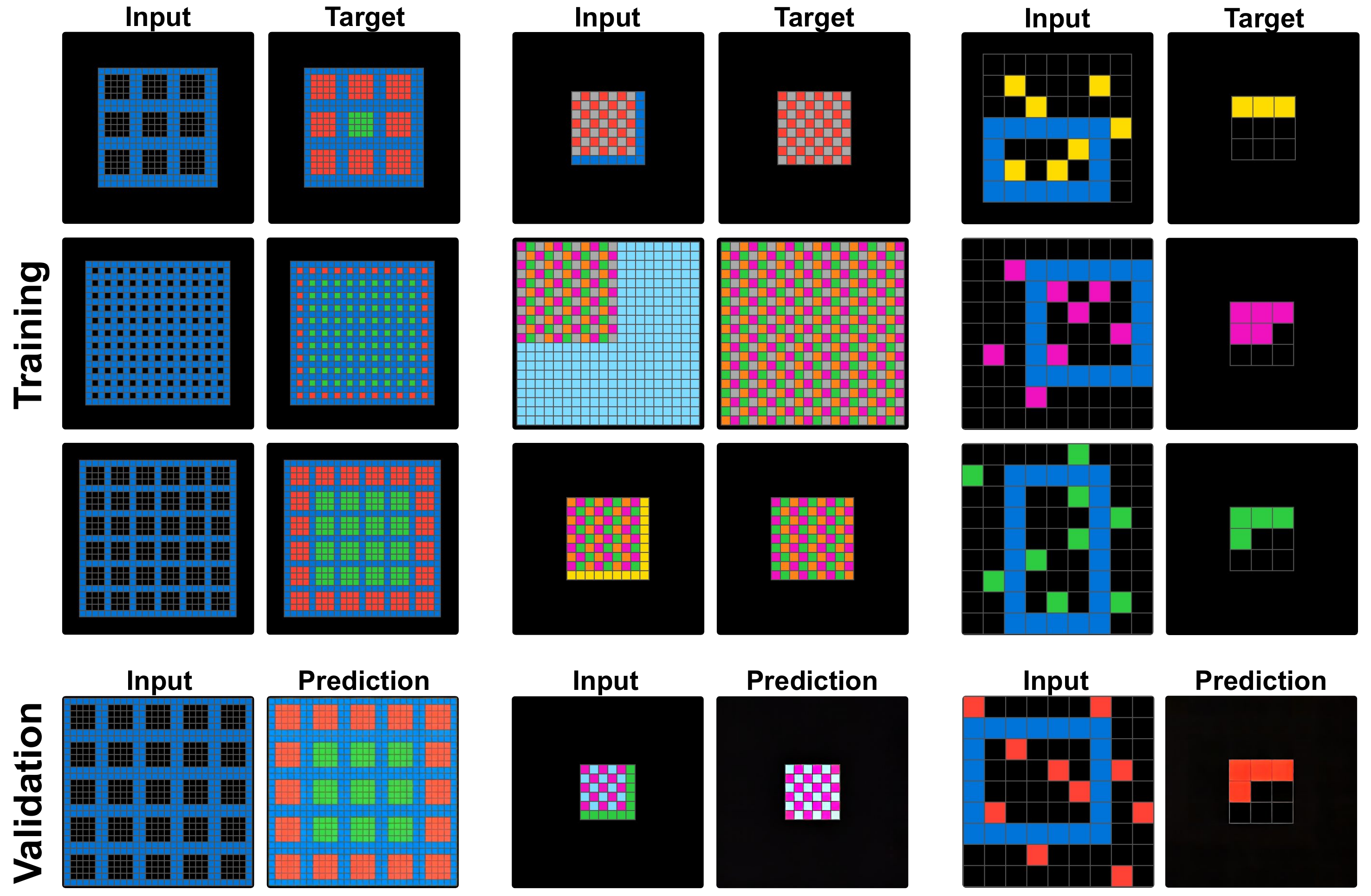}
  \caption{Additional examples of solved tasks from ARC-AGI. Training samples for each task (first three rows), followed by evaluation (last row).}
  \label{appdx:fig:arc-1}
\end{figure}

\begin{figure}[htbp]
  \centering
  \includegraphics[width=1.0\linewidth]{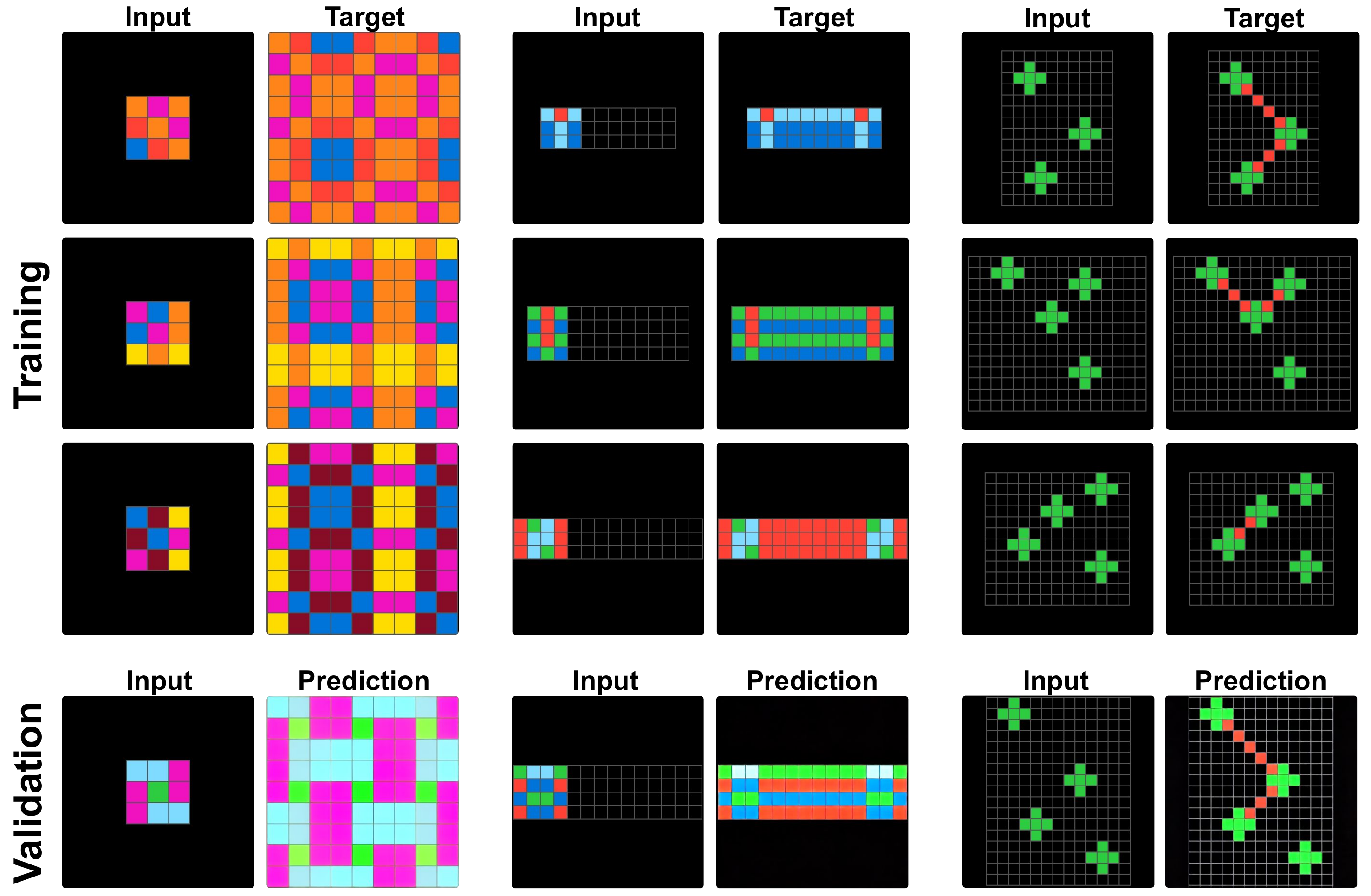}
  \caption{Additional examples of solved tasks from ARC-AGI. Training samples for each task (first three rows), followed by evaluation (last row).}
  \label{appdx:fig:arc-2}
\end{figure}

Finally, we show some additional results for the ARC-AGI task in Figures~\ref{appdx:fig:arc-0}, \ref{appdx:fig:arc-1} and \ref{appdx:fig:arc-2}.

\section{Additional Limitations}
Here we briefly outline further limitations of our approach. Some are intrinsic to the few-shot learning paradigm, while others stem from specific choices and constraints in our current implementation.

\paragraph{Uncertainty in Task Definitions.}
Few-shot tasks, especially in the one-shot setting, are often underspecified, which introduces significant ambiguity. A single example is rarely sufficient to capture the full variability of a class or to clearly convey the intended concept, particularly in the absence of strong inductive priors.

In our setting, even if the model \textbf{has access to the necessary representational features} required to support a generalizable solution, there is no guarantee that the LoRA adaptation will make effective use of them. We believe that the model often relies on these features when they provide a low-effort solution to the task, suggesting that these features serve as strong priors for the model. However, in some cases, especially when only a single example is available, the model may instead exploit superficial correlations or shortcuts that produce the correct output without generalization.

\begin{figure}[htbp]
  \centering
  \includegraphics[width=0.8\linewidth]{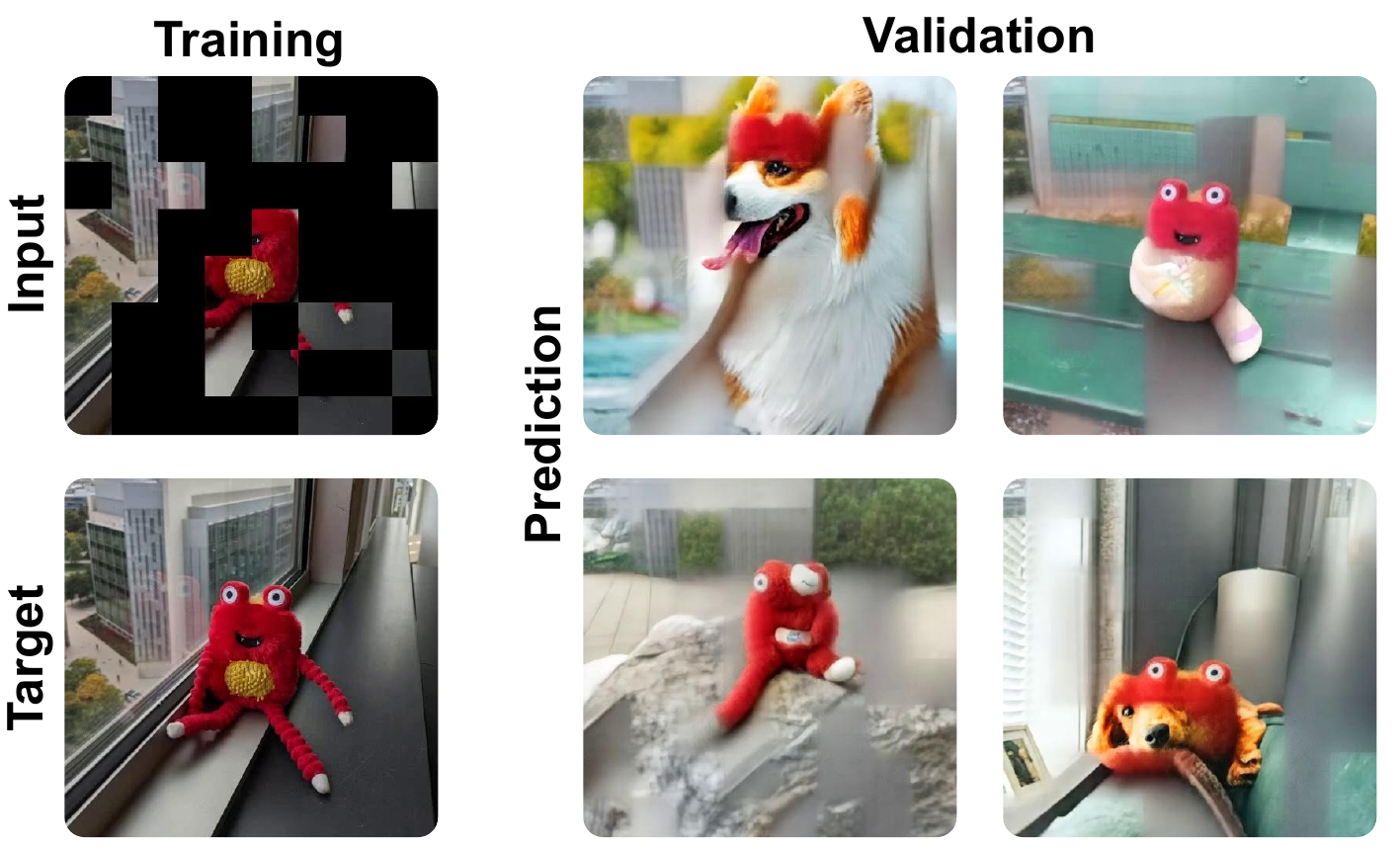}
    \caption{Left: a training sample used in the inpainting task (one of several masked-region variations). Right: representative validation outputs, where the model erroneously inserts features reminiscent of the original image.}
  \label{appdx:fig:limitations/inpainting}
\end{figure}

\begin{figure}[htbp]
  \centering
  \includegraphics[width=0.8\linewidth]{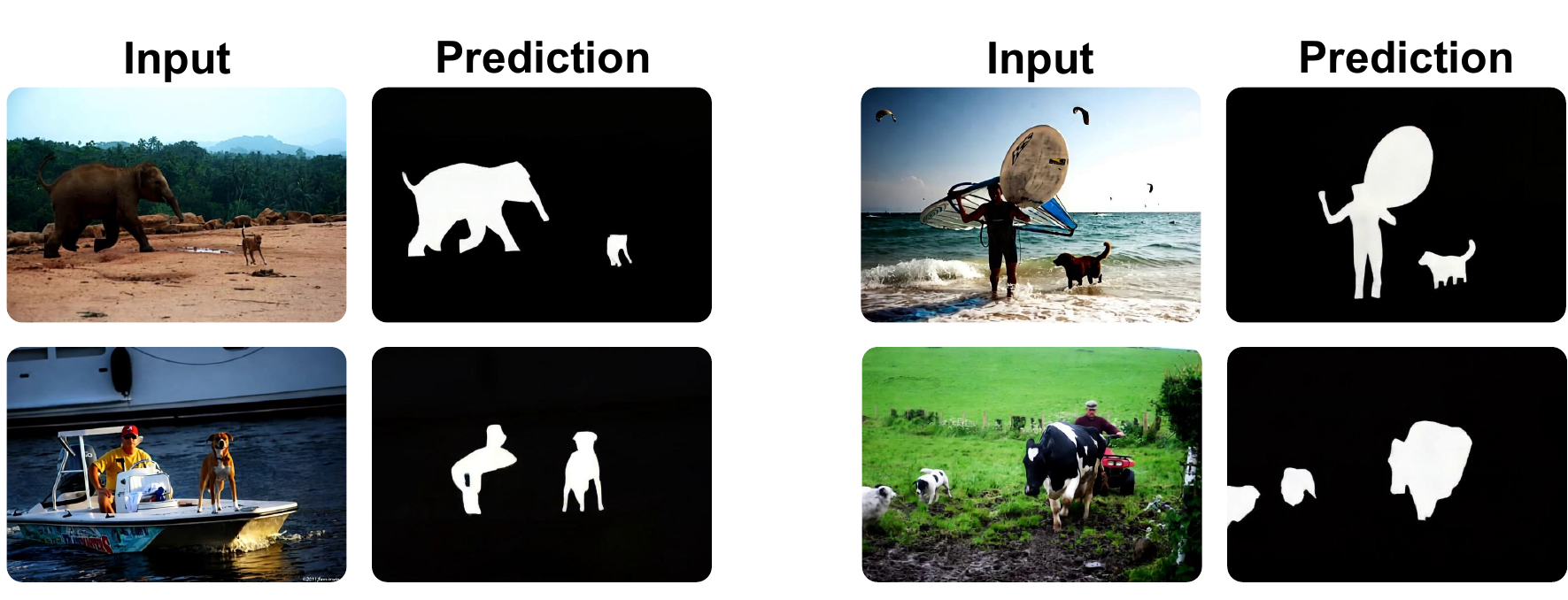}
    \caption{Examples from the binary segmentation task (validation), where the model, trained to segment dogs, instead segments broader categories like all animals or defaults to spatial biases such as centered objects.}
  \label{appdx:fig:limitations/binary-seg}
\end{figure}

As shown in Figure~\ref{appdx:fig:limitations/inpainting}, a model trained for inpainting on a single image can, in some cases, introduce features into validation examples that reflect artifacts of the original training task. A related pattern appears in the binary segmentation setting, where the model may adopt broader interpretations such as "segment animals" instead of the more specific "segment dogs," or even default to "segment centered object" when trained with limited data (see Figure~\ref{appdx:fig:limitations/binary-seg}). These behaviors illustrate how underspecification in few-shot scenarios can lead to outputs that diverge from the intended outcomes.

\begin{figure}[htbp]
  \centering
  \includegraphics[width=0.8\linewidth]{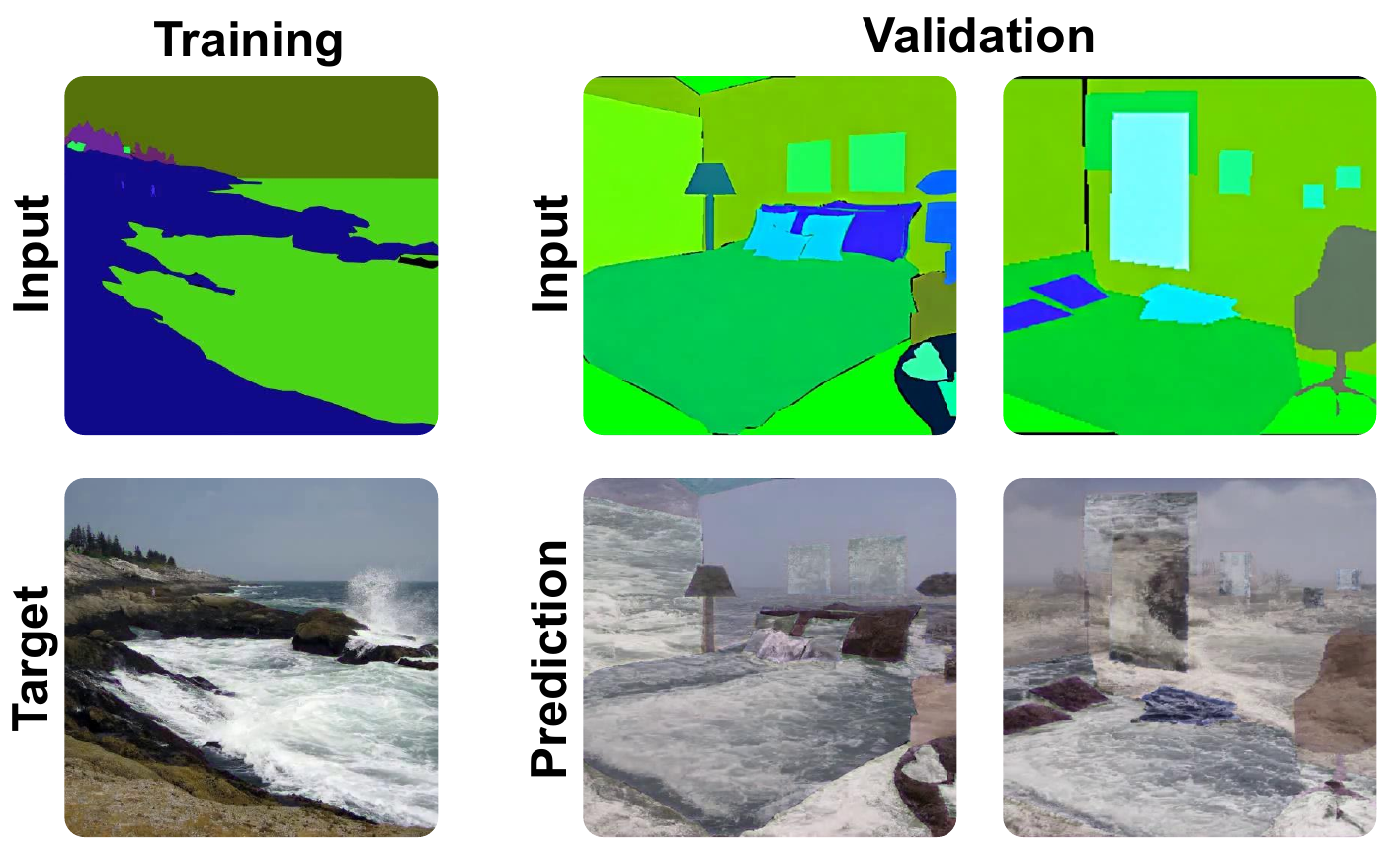}
    \caption{Examples for Segmentation to Image with out-of-distribution validation samples.}
  \label{appdx:fig:limitations/limitations-seg2mask-ood}
\end{figure}

An extreme example of this is shown in Figure~\ref{appdx:fig:limitations/limitations-seg2mask-ood}, where the model clearly learns to texturize the segmentations, which is a completely sensible outcome, rather than deriving any semantic understanding from the shapes to construct the images; consequently, it does not handle out-of-distribution images.

\paragraph{Limitations of RGB Representations.}
Tasks must be encoded and decoded into and from RGB space. This poses limitations for tasks that require different modalities that cannot be easily encoded into RGB. Furthermore, tasks requiring high  RGB-accuracy representations may be harder to tackle in a few-shot setting as there can be color shifts. Nevertheless, we have observed a similar color shift in \textbf{both} training and test samples, suggesting that one could explore building a deterministic mapping from the training data to tackle this problem.  

\begin{figure}[htbp]
  \centering
  \includegraphics[width=0.9\linewidth]{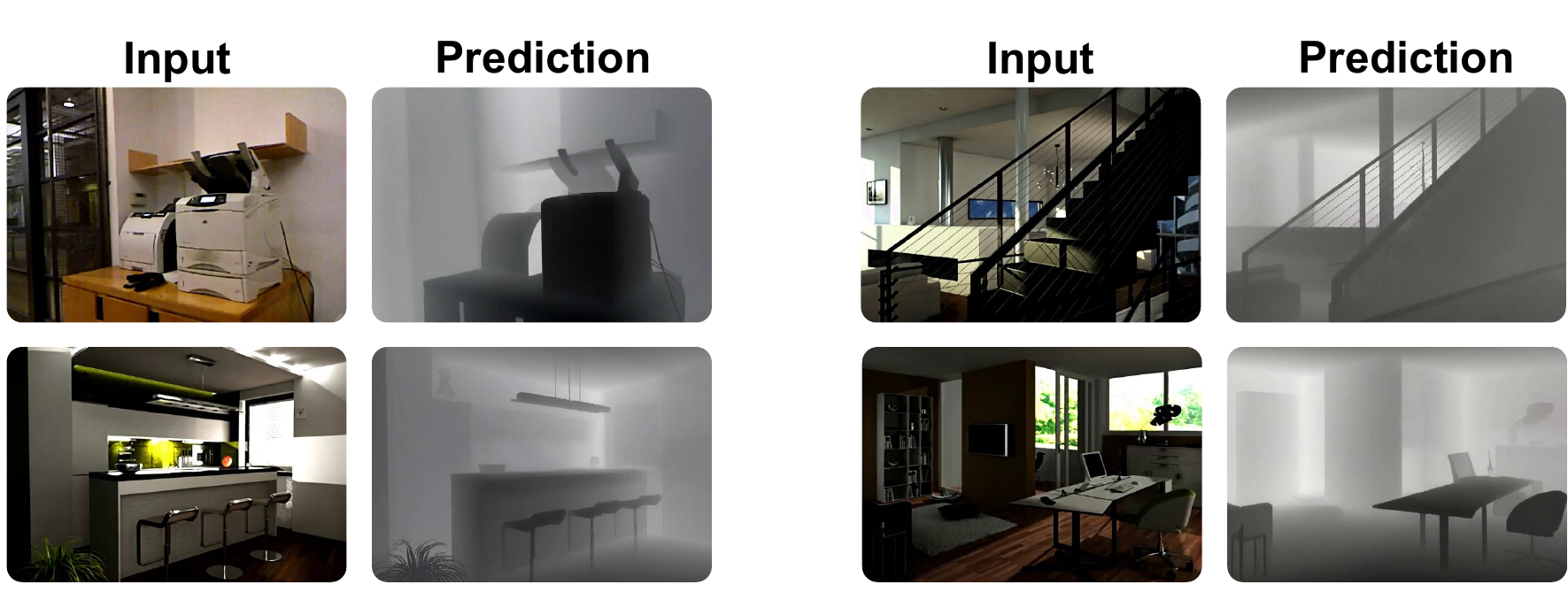}
    \caption{Validation examples for depth estimation.}
  \label{appdx:fig:limitations/limitations-depth}
\end{figure}

This can occur, for example, in depth estimation tasks, where accurate encoding and decoding of RGB values is crucial for precise depth prediction. We illustrate such examples in Figure \ref{appdx:fig:limitations/limitations-depth}, even though the images seem to give reasonable depth approximations, they do not provide good depth estimations.

\section{Broader Impact Considerations}

This work investigates the internal representations learned by Video Diffusion Models, with the aim of better understanding their structure and capabilities. It does not introduce new generative techniques or applications, and we do not anticipate direct societal consequences arising from our contributions.

However, we recognize that foundational research can influence the development and deployment of future technologies. By improving interpretability and adaptability of generative models, our work may inform subsequent applications. As such models become more widely used, understanding their internal mechanisms is an important step toward ensuring they are applied in safe, fair, and responsible ways. While we do not identify specific risks associated with our method, we believe careful analysis and transparency are essential in supporting the broader goals of ethical machine learning research.

%% file: paper.bbl
\begin{thebibliography}{10}

\bibitem{akyurek2025surprisingeffectivenesstesttimetraining}
Ekin Akyürek, Mehul Damani, Adam Zweiger, Linlu Qiu, Han Guo, Jyothish Pari, Yoon Kim, and Jacob Andreas.
\newblock The surprising effectiveness of test-time training for few-shot learning.
\newblock {\em arXiv preprint arXiv:2411.07279}, 2025.

\bibitem{bai2024sequential}
Yutong Bai, Xinyang Geng, Karttikeya Mangalam, Amir Bar, Alan~L Yuille, Trevor Darrell, Jitendra Malik, and Alexei~A Efros.
\newblock Sequential modeling enables scalable learning for large vision models.
\newblock In {\em Proceedings of the IEEE/CVF Conference on Computer Vision and Pattern Recognition}, pages 22861--22872, 2024.

\bibitem{bar2022visual}
Amir Bar, Yossi Gandelsman, Trevor Darrell, Amir Globerson, and Alexei~A. Efros.
\newblock Visual prompting via image inpainting.
\newblock {\em CoRR}, abs/2209.00647, 2022.

\bibitem{berman2024record}
Jeremy Berman.
\newblock How {I} got a record 53.6\% on {ARC-AGI-Pub} using {Sonnet} 3.5.1.
\newblock \url{https://jeremyberman.substack.com/p/how-i-got-a-record-536-on-arc-agi}, 2024.
\newblock Accessed: 2025-05-12.

\bibitem{blattmann2023stablevideodiffusionscaling}
Andreas Blattmann, Tim Dockhorn, Sumith Kulal, Daniel Mendelevitch, Maciej Kilian, Dominik Lorenz, Yam Levi, Zion English, Vikram Voleti, Adam Letts, Varun Jampani, and Robin Rombach.
\newblock Stable video diffusion: Scaling latent video diffusion models to large datasets.
\newblock {\em CoRR}, abs/2311.15127, 2023.

\bibitem{chollet2019measure}
Fran{\c{c}}ois Chollet.
\newblock On the measure of intelligence.
\newblock {\em arXiv preprint arXiv:1911.01547}, 2019.

\bibitem{chollet2024arc}
Francois Chollet, Mike Knoop, Gregory Kamradt, and Bryan Landers.
\newblock Arc prize 2024: Technical report.
\newblock {\em arXiv preprint arXiv:2412.04604}, 2024.

\bibitem{delatolas2025studying}
Thanos Delatolas, Vicky Kalogeiton, and Dim~P Papadopoulos.
\newblock Studying image diffusion features for zero-shot video object segmentation.
\newblock {\em arXiv preprint arXiv:2504.05468}, 2025.

\bibitem{dong2024surveyincontextlearning}
Qingxiu Dong, Lei Li, Damai Dai, Ce~Zheng, Jingyuan Ma, Rui Li, Heming Xia, Jingjing Xu, Zhiyong Wu, Baobao Chang, Xu~Sun, and Zhifang Sui.
\newblock A survey on in-context learning.
\newblock In {\em EMNLP}, pages 1107--1128, 2024.

\bibitem{polyak2025moviegencastmedia}
Adam~Polyak et~al.
\newblock Movie gen: A cast of media foundation models.
\newblock {\em CoRR}, abs/2410.13720, 2024.

\bibitem{deepseekr1_2025}
DeepSeek-AI et~al.
\newblock Deepseek-r1: Incentivizing reasoning capability in llms via reinforcement learning.
\newblock {\em CoRR}, abs/2501.12948, January 2025.

\bibitem{geminiteam2024gemini15}
Machel~Reid et~al.
\newblock Gemini 1.5: Unlocking multimodal understanding across millions of tokens of context.
\newblock {\em CoRR}, abs/2403.05530, 2024.

\bibitem{agarwal2025cosmos}
Niket~Agarwal et~al.
\newblock Cosmos world foundation model platform for physical ai.
\newblock {\em CoRR}, abs/2501.03575, January 2025.

\bibitem{villegas2022phenakivariablelengthvideo}
Ruben~Villegas et~al.
\newblock Phenaki: Variable length video generation from open domain textual description.
\newblock {\em CoRR}, abs/2210.02399, 2022.

\bibitem{brown2020language}
Tom B.~Brown et~al.
\newblock Language models are few-shot learners.
\newblock {\em CoRR}, abs/2005.14165, 2020.

\bibitem{hacohen2024ltx}
Yoav~HaCohen et~al.
\newblock Ltx-video: Realtime video latent diffusion.
\newblock {\em CoRR}, abs/2501.00103, January 2025.

\bibitem{yang2024cogvideox}
Zhuoyi~Yang et~al.
\newblock Cogvideox: Text-to-video diffusion models with an expert transformer.
\newblock {\em CoRR}, abs/2408.06072, 2024.

\bibitem{feynman1989}
Richard~P. Feynman.
\newblock Feynman's office; the last blackboards.
\newblock {\em Physics Today}, 42(2):88--88, February 1989.

\bibitem{friston2010free}
Karl Friston.
\newblock The free-energy principle: a unified brain theory?
\newblock {\em Nature reviews neuroscience}, 11(2):127--138, 2010.

\bibitem{geng2024instructdiffusion}
Zigang Geng, Binxin Yang, Tiankai Hang, Chen Li, Shuyang Gu, Ting Zhang, Jianmin Bao, Zheng Zhang, Houqiang Li, Han Hu, et~al.
\newblock Instructdiffusion: A generalist modeling interface for vision tasks.
\newblock In {\em Proceedings of the IEEE/CVF Conference on computer vision and pattern recognition}, pages 12709--12720, 2024.

\bibitem{giannone2022few}
Giorgio Giannone, Didrik Nielsen, and Ole Winther.
\newblock Few-shot diffusion models.
\newblock {\em CoRR}, abs/2205.15463, 2022.

\bibitem{deepmind2024veo2}
{Google DeepMind}.
\newblock Veo 2.
\newblock \url{https://deepmind.google/technologies/veo/veo-2/}, December 2024.
\newblock Accessed: 2025-05-12.

\bibitem{grattafiori2024llama}
Aaron Grattafiori, Abhimanyu Dubey, Abhinav Jauhri, Abhinav Pandey, Abhishek Kadian, Ahmad Al-Dahle, Aiesha Letman, Akhil Mathur, Alan Schelten, Alex Vaughan, et~al.
\newblock The llama 3 herd of models.
\newblock {\em arXiv e-prints}, pages arXiv--2407, 2024.

\bibitem{hassan2024gemgeneralizableegovisionmultimodal}
Mariam Hassan, Sebastian Stapf, Ahmad Rahimi, Pedro M~B Rezende, Yasaman Haghighi, David Brüggemann, Isinsu Katircioglu, Lin Zhang, Xiaoran Chen, Suman Saha, Marco Cannici, Elie Aljalbout, Botao Ye, Xi~Wang, Aram Davtyan, Mathieu Salzmann, Davide Scaramuzza, Marc Pollefeys, Paolo Favaro, and Alexandre Alahi.
\newblock Gem: A generalizable ego-vision multimodal world model for fine-grained ego-motion, object dynamics, and scene composition control.
\newblock {\em CVPR}, 2025.

\bibitem{helbling2025conceptattention}
Alec Helbling, Tuna Han~Salih Meral, Benjamin Hoover, Pinar Yanardag, and Duen~Horng Chau.
\newblock Conceptattention: Diffusion transformers learn highly interpretable features.
\newblock {\em CoRR}, abs/2502.04320, February 2025.

\bibitem{ho2020denoising}
Jonathan Ho, Ajay Jain, and Pieter Abbeel.
\newblock Denoising diffusion probabilistic models.
\newblock {\em Advances in neural information processing systems}, 33:6840--6851, 2020.

\bibitem{hong2022cogvideolargescalepretrainingtexttovideo}
Wenyi Hong, Ming Ding, Wendi Zheng, Xinghan Liu, and Jie Tang.
\newblock Cogvideo: Large-scale pretraining for text-to-video generation via transformers.
\newblock {\em CoRR}, abs/2205.15868, 2022.

\bibitem{hu2021loralowrankadaptationlarge}
Edward~J Hu, yelong shen, Phillip Wallis, Zeyuan Allen-Zhu, Yuanzhi Li, Shean Wang, Lu~Wang, and Weizhu Chen.
\newblock Lo{RA}: Low-rank adaptation of large language models.
\newblock In {\em International Conference on Learning Representations}, 2022.

\bibitem{huang2024lorahub}
Chengsong Huang, Qian Liu, Bill~Yuchen Lin, Tianyu Pang, Chao Du, and Min Lin.
\newblock Lorahub: Efficient cross-task generalization via dynamic lora composition.
\newblock {\em arXiv preprint arXiv:2307.13269}, 2024.

\bibitem{kanervisto2025world}
Anssi Kanervisto, Dave Bignell, Linda~Yilin Wen, Martin Grayson, Raluca Georgescu, Sergio Valcarcel~Macua, Shan~Zheng Tan, Tabish Rashid, Tim Pearce, Yuhan Cao, et~al.
\newblock World and human action models towards gameplay ideation.
\newblock {\em Nature}, 638(8051):656--663, 2025.

\bibitem{flux2024}
Black~Forest Labs.
\newblock Flux.1-dev.
\newblock \url{https://huggingface.co/black-forest-labs/FLUX.1-dev}, 2025.

\bibitem{lin2014microsoft}
Tsung-Yi Lin, Michael Maire, Serge~J. Belongie, Lubomir~D. Bourdev, Ross~B. Girshick, James Hays, Pietro Perona, Deva Ramanan, Piotr Dollár, and C.~Lawrence Zitnick.
\newblock Microsoft coco: Common objects in context.
\newblock {\em CoRR}, abs/1405.0312, 2014.

\bibitem{lin2025realgeneral}
Yijing Lin, Mengqi Huang, Shuhan Zhuang, and Zhendong Mao.
\newblock Realgeneral: Unifying visual generation via temporal in-context learning with video models.
\newblock {\em arXiv preprint arXiv:2503.10406}, 2025.

\bibitem{liu2022fewshotparameterefficientfinetuningbetter}
Haokun Liu, Derek Tam, Mohammed Muqeeth, Jay Mohta, Tenghao Huang, Mohit Bansal, and Colin Raffel.
\newblock Few-shot parameter-efficient fine-tuning is better and cheaper than in-context learning.
\newblock {\em CoRR}, abs/2205.05638, 2022.

\bibitem{liu2024ada}
Jia Liu, Changlin Li, Qirui Sun, Jiahui Ming, Chen Fang, Jue Wang, Bing Zeng, and Shuaicheng Liu.
\newblock Ada-adapter:fast few-shot style personlization of diffusion model with pre-trained image encoder.
\newblock {\em CoRR}, abs/2407.05552, 2024.

\bibitem{tiny-imagenet}
mnmoustafa and Mohammed Ali.
\newblock Tiny imagenet.
\newblock \url{https://kaggle.com/competitions/tiny-imagenet}, 2017.
\newblock Kaggle.

\bibitem{moskvichev2023conceptarc}
Arsenii Moskvichev, Victor~Vikram Odouard, and Melanie Mitchell.
\newblock The conceptarc benchmark: Evaluating understanding and generalization in the arc domain.
\newblock {\em Trans. Mach. Learn. Res.}, 2023, 2023.

\bibitem{najafi2024rifflearningrephraseinputs}
Saeed Najafi and Alona Fyshe.
\newblock Riff: Learning to rephrase inputs for few-shot fine-tuning of language models.
\newblock {\em arXiv preprint arXiv:2403.02271}, 2024.

\bibitem{nari2025dia}
{Nari Labs}.
\newblock Dia: A tts model capable of generating ultra-realistic dialogue in one pass.
\newblock \url{https://github.com/nari-labs/dia}, 2025.
\newblock Accessed: 2025-05-12.

\bibitem{Silberman:ECCV12}
Pushmeet~Kohli Nathan~Silberman, Derek~Hoiem and Rob Fergus.
\newblock Indoor segmentation and support inference from rgbd images.
\newblock In {\em ECCV}, 2012.

\bibitem{achiam2023gpt}
OpenAI.
\newblock {GPT-4} technical report.
\newblock {\em arXiv preprint arXiv:2303.08774}, 2023.

\bibitem{parr2022active}
Thomas Parr, Giovanni Pezzulo, and Karl~J Friston.
\newblock {\em Active inference: the free energy principle in mind, brain, and behavior}.
\newblock MIT Press, 2022.

\bibitem{prabhakar2024lorasoups}
Akshara Prabhakar, Yuanzhi Li, Karthik Narasimhan, Sham~M. Kakade, Eran Malach, and Samy Jelassi.
\newblock Lora soups: Merging loras for practical skill composition tasks.
\newblock {\em CoRR}, abs/2410.13025, 2024.

\bibitem{brooks2024video}
Yiran Qin, Zhelun Shi, Jiwen Yu, Xijun Wang, Enshen Zhou, Lijun Li, Zhenfei Yin, Xihui Liu, Lu~Sheng, Jing Shao, Lei Bai, Wanli Ouyang, and Ruimao Zhang.
\newblock Worldsimbench: Towards video generation models as world simulators.
\newblock {\em CoRR}, abs/2410.18072, 2024.

\bibitem{roberts:2021}
Mike Roberts, Jason Ramapuram, Anurag Ranjan, Atulit Kumar, Miguel~Angel Bautista, Nathan Paczan, Russ Webb, and Joshua~M. Susskind.
\newblock {Hypersim}: {A} photorealistic synthetic dataset for holistic indoor scene understanding.
\newblock In {\em International Conference on Computer Vision (ICCV) 2021}, 2021.

\bibitem{rombach2022highresolutionimagesynthesislatent}
Robin Rombach, Andreas Blattmann, Dominik Lorenz, Patrick Esser, and Bj{\"o}rn Ommer.
\newblock High-resolution image synthesis with latent diffusion models.
\newblock In {\em Proceedings of the IEEE/CVF conference on computer vision and pattern recognition}, pages 10684--10695, 2022.

\bibitem{ruiz2023dreambooth}
Nataniel Ruiz, Yuanzhen Li, Varun Jampani, Yael Pritch, Michael Rubinstein, and Kfir Aberman.
\newblock Dreambooth: Fine tuning text-to-image diffusion models for subject-driven generation.
\newblock {\em CoRR}, abs/2208.12242, 2022.

\bibitem{sesame_uncanny_valley_voice}
{Sesame}.
\newblock Crossing the uncanny valley of voice.
\newblock \url{https://www.sesame.com/research/crossing_the_uncanny_valley_of_voice}, n.d.
\newblock Accessed: 2025-05-12.

\bibitem{tang2023emergent}
Luming Tang, Menglin Jia, Qianqian Wang, Cheng~Perng Phoo, and Bharath Hariharan.
\newblock Emergent correspondence from image diffusion.
\newblock {\em Advances in Neural Information Processing Systems}, 36:1363--1389, 2023.

\bibitem{Vaswani2017AttentionIA}
Ashish Vaswani, Noam Shazeer, Niki Parmar, Jakob Uszkoreit, Llion Jones, Aidan~N. Gomez, Lukasz Kaiser, and Illia Polosukhin.
\newblock Attention is all you need.
\newblock In {\em NIPS}, pages 6000--6010, 2017.

\bibitem{wang2024zero}
Qian Wang, Abdelrahman Eldesokey, Mohit Mendiratta, Fangneng Zhan, Adam Kortylewski, Christian Theobalt, and Peter Wonka.
\newblock Zero-shot video semantic segmentation based on pre-trained diffusion models.
\newblock {\em CoRR}, abs/2405.16947, 2024.

\bibitem{wang2023images}
Xinlong Wang, Wen Wang, Yue Cao, Chunhua Shen, and Tiejun Huang.
\newblock Images speak in images: A generalist painter for in-context visual learning.
\newblock In {\em Proceedings of the IEEE/CVF Conference on Computer Vision and Pattern Recognition}, pages 6830--6839, 2023.

\bibitem{wang2023context}
Zhendong Wang, Yifan Jiang, Yadong Lu, Pengcheng He, Weizhu Chen, Zhangyang Wang, Mingyuan Zhou, et~al.
\newblock In-context learning unlocked for diffusion models.
\newblock {\em Advances in Neural Information Processing Systems}, 36:8542--8562, 2023.

\bibitem{llm_emergent}
Jason Wei, Yi~Tay, Rishi Bommasani, Colin Raffel, Barret Zoph, Sebastian Borgeaud, Dani Yogatama, Maarten Bosma, Denny Zhou, Donald Metzler, Ed~H. Chi, Tatsunori Hashimoto, Oriol Vinyals, Percy Liang, Jeff Dean, and William Fedus.
\newblock Emergent abilities of large language models.
\newblock {\em CoRR}, abs/2206.07682, 2022.

\bibitem{xiao2024omnigen}
Shitao Xiao, Yueze Wang, Junjie Zhou, Huaying Yuan, Xingrun Xing, Ruiran Yan, Shuting Wang, Tiejun Huang, and Zheng Liu.
\newblock Omnigen: Unified image generation.
\newblock {\em CoRR}, abs/2409.11340, 2024.

\bibitem{xu2024matters}
Guangkai Xu, Yongtao Ge, Mingyu Liu, Chengxiang Fan, Kangyang Xie, Zhiyue Zhao, Hao Chen, and Chunhua Shen.
\newblock What matters when repurposing diffusion models for general dense perception tasks?
\newblock {\em The Thirteenth International Conference on Learning Representations (ICLR)}, 2025.

\bibitem{yang2024video}
Sherry Yang, Jacob~C. Walker, Jack Parker-Holder, Yilun Du, Jake Bruce, André Barreto, Pieter Abbeel, and Dale Schuurmans.
\newblock Video as the new language for real-world decision making.
\newblock {\em CoRR}, abs/2402.17139, 2024.

\bibitem{zheng2024opensorademocratizingefficientvideo}
Zangwei Zheng, Xiangyu Peng, Tianji Yang, Chenhui Shen, Shenggui Li, Hongxin Liu, Yukun Zhou, Tianyi Li, and Yang You.
\newblock Open-sora: Democratizing efficient video production for all.
\newblock {\em arXiv preprint arXiv:2412.20404}, 2024.

\bibitem{zhou2017scene}
Bolei Zhou, Hang Zhao, Xavier Puig, Sanja Fidler, Adela Barriuso, and Antonio Torralba.
\newblock Scene parsing through ade20k dataset.
\newblock In {\em Proceedings of the IEEE Conference on Computer Vision and Pattern Recognition}, 2017.

\bibitem{zhou2019semantic}
Bolei Zhou, Hang Zhao, Xavier Puig, Tete Xiao, Sanja Fidler, Adela Barriuso, and Antonio Torralba.
\newblock Semantic understanding of scenes through the ade20k dataset.
\newblock {\em International Journal of Computer Vision}, 127(3):302--321, 2019.

\end{thebibliography}
